\documentclass[11pt]{article}

\usepackage[margin=1in]{geometry}
\usepackage[T1]{fontenc}
\usepackage[utf8]{inputenc}
\usepackage{lmodern}
\usepackage{microtype}
\usepackage{amsmath,amssymb,amsfonts,bm,mathtools}
\usepackage{booktabs,longtable,array,multirow}
\usepackage{graphicx}
\usepackage{float}
\usepackage{xcolor}
\usepackage{hyperref}
\usepackage{enumitem}
\usepackage{caption}
\usepackage{algorithm}
\usepackage{algpseudocode}
\usepackage[numbers,sort&compress]{natbib}
\usepackage{authblk}

\hypersetup{
  colorlinks=true,
  linkcolor=blue!50!black,
  citecolor=blue!50!black,
  urlcolor=blue!50!black
}

\newcommand{\cfgfield}[1]{\nolinkurl{#1}}

\title{\textbf{A Synthetic Multivariate Refrigerator Time-Series Dataset for Predictive Maintenance}}

\author{Islam Benamirouche}
\author{Feriel Fass}
\author{Djemel Ziou\thanks{Corresponding author: \href{mailto:djemel.ziou@usherbrooke.ca}{djemel.ziou@usherbrooke.ca}}}
\affil{D\'epartement d'informatique, Universit\'e de Sherbrooke, Sherbrooke, QC, Canada}
\date{}

\begin{document}
\maketitle

\begin{abstract}
We generated synthetic multivariate time series for 27 refrigerators
with a simplified physics-inspired simulator at one-minute resolution.
The simulator includes ambient-temperature variation, door use,
thermostat and compressor operation, heat exchange, defrost, electrical
consumption, and six progressive degradation types. Each refrigerator
provides 15 to 20 sensor outputs according to its configuration. The
dataset contains 7,066,161 rows in 27 time-series files and 27 failure
logs. The release also includes the Python generator, refrigerator
configurations, and documentation. The data can support failure
prediction, degradation analysis, and learning across refrigerators with
different sensor-output sets.
\end{abstract}

\section{Background and Summary}

Refrigerators operate continuously and depend on coupled thermal,
electrical, mechanical, and control processes. Their operating condition
can be characterized using temperatures, compressor current, power,
vibration, humidity, pressure, and other sensor outputs. Reliable
refrigerator operation is important for food preservation and energy
efficiency~\cite{ASHRAE2022,Hueppe2021,deFrias2020}. Predictive
maintenance additionally aims to identify developing failures before
they interrupt normal operation. However, publicly accessible
refrigerator datasets providing continuous multivariate measurements
together with explicit degradation-start and failure-interval
annotations remain limited.

Collecting failure data for predictive maintenance is generally
difficult because failure observations are scarce, datasets are often
imbalanced, and run-to-failure data are costly to obtain
\cite{Nieminen2026}. Refrigerator behaviour also varies with equipment
age, operating conditions, usage patterns, and maintenance history.
Sensor configurations differ, with some
refrigerators providing only common temperature and control variables
and others including current, humidity, pressure, vibration, acoustic
noise, frost thickness, or performance-related outputs. In addition, the
start of a slow degradation is usually not known in real refrigerator
data.

Fonseca et al.~\cite{fonseca2023homeappliancedataset} provide current and
vibration measurements collected from home appliances, including
refrigerators, in a repair-center setting. Their data are organized as
extracted operating segments rather than uninterrupted monitoring over
several months and therefore do not describe multiday sensor-output
changes from degradation start to failure occurrence. Simulation
provides an alternative when real degradation histories are
unavailable; for example, Polyakov et
al.~\cite{Polyakov2025DigitalTwinCooling} generated synthetic data for
condenser anomalies in a data-center cooling system.
Simplified and grey-box refrigerator models have been developed to
describe thermal and control dynamics
\cite{Tagliafico2012,Sossan2016}, while more detailed dynamic
refrigeration-cycle models have also been validated experimentally
\cite{Caglayan2022}.

Compared with these refrigerator-related datasets and simulation
studies, the dataset presented here provides continuous
multivariate time series at one-minute resolution, explicit degradation
start and failure occurrence times, multiple progressive degradation
types, and different sensor-output configurations across a family of
simulated refrigerators.

The remainder of the paper presents the dataset specifications in
Section~\ref{overview}, the generation method in
Section~\ref{sec:methods}, and the technical validation in
Section~\ref{sec:technical_validation}.

\section{Dataset Specifications}
\label{overview}

\begin{longtable}{>{\raggedright\arraybackslash}p{0.27\linewidth}
                  >{\raggedright\arraybackslash}p{0.65\linewidth}}
\caption{Specifications of the released dataset.}
\label{tab:dataset-specifications}\\
\toprule
Specification & Description \\
\midrule
\endfirsthead
\multicolumn{2}{c}{\tablename\ \thetable\ (continued)}\\
\toprule
Specification & Description \\
\midrule
\endhead
\bottomrule
\endfoot

Subject &
Computer Science \\

Specific subject area &
Synthetic multivariate time series for refrigerator predictive
maintenance and failure prediction \\

Type of data &
Minute-resolution sensor time series, failure logs in CSV format, refrigerator configurations, generation code, and documentation \\

Data collection &
Twenty-seven refrigerators were simulated independently at one-minute
resolution using a simplified physics-inspired simulator. Each
refrigerator provides 15 to 20 sensor outputs and includes six
degradation types. \\

Data accessibility &
The dataset is publicly available on Zenodo
\cite{BenamiroucheDataset2026}. Generation code,
refrigerator configurations, and documentation are available through
the GitHub repository described in the Code Availability section. \\

Related research article &
Benamirouche et al., \textit{A Transferable Autologistic Model for
Predicting Rare Failures in Heterogeneous Equipment},
arXiv:2608.06695~\cite{benamirouche2026transferable} \\

\end{longtable}

\section{Data Generation Method}
\label{sec:methods}

Each refrigerator was simulated independently with a sampling
interval \(\Delta t=1\) min. For refrigerator \(e\),
\(t\in\{1,\ldots,T_e\}\) indexes the one-minute observations, and
\(T_e\) is the number of generated observations. Each value of \(t\)
corresponds to one row of the released time-series table.
For refrigerator \(e\), \(\theta_e\) denotes the fixed configuration.
Before the minute-level simulation begins, the simulator uses it to
construct the failure log \(\mathcal F_e\), the planned defrost schedule
\(\mathcal D_e\), and the initial internal state \(Z_{e,0}\).
The quantities \(\mathcal D_e\) and \(Z_{e,t}\) are internal to the
simulator.

Section~\ref{app:failure-scheduling} of the Appendix describes failure scheduling and
degradation timing. Section~\ref{app:phenomena} of the Appendix gives the
phenomenon simulation rules, and
Section~\ref{app:output-assembly} of the Appendix summarizes the released sensor and
annotation fields.

\subsection{Simulator Inputs and Generated Data}
\label{sec:generator-inputs}

The entries of \(\theta_e\) remain unchanged during
\(t=1,\ldots,T_e\) and include initial and target temperatures,
thermostat thresholds, ambient-forcing parameters, door-opening
probabilities, defrost timing, reference electrical values, failure-scheduling
settings, and optional-sensor flags.

At time instant \(t\), refrigerator \(e\) emits the sensor-output
vector
\begin{equation}
\label{eq:sensor-output-vector}
X_t^{(e)}
=
\left(
X_{t,1}^{(e)},
X_{t,2}^{(e)},
\ldots,
X_{t,m_e}^{(e)}
\right)^\top
\in\mathbb R^{m_e},
\end{equation}
where \(X_{t,i}^{(e)}\) is the \(i\)-th sensor output and \(m_e\) is
the number of sensor outputs available for refrigerator \(e\). The
quantity \(m_e\) denotes the number of emitted variables and not the
number of physical sensors.

The multivariate time series of refrigerator \(e\) is
\begin{equation}
\label{eq:sensor-output-series}
\mathbf X^{(e)}
=
\begin{pmatrix}
\left(X_1^{(e)}\right)^\top\\
\left(X_2^{(e)}\right)^\top\\
\vdots\\
\left(X_{T_e}^{(e)}\right)^\top
\end{pmatrix}
\in\mathbb R^{T_e\times m_e}.
\end{equation}
Each row corresponds to one discrete time instant and each column
contains the time series of one sensor output.

The failure log of refrigerator \(e\) is written as
\begin{equation}
\label{eq:failure-log}
\mathcal F_e
=
\left(
F_{e,1},
F_{e,2},
\ldots,
F_{e,M_e}
\right),
\end{equation}
where \(M_e=\lvert\mathcal F_e\rvert\) is the number of retained failures
and \(F_{e,j}\) is the \(j\)-th retained failure.

The binary label \(y_{e,t}\) indicates whether refrigerator \(e\) is
operating or is within a failure interval at time \(t\). It is derived
from the failure log and is defined by
\begin{equation}
\label{eq:equipment-state-label}
y_{e,t}
=
\begin{cases}
1,
& \text{if there exists \(j\) such that }
t_{e,j}^{\mathrm{fail}}
\le t
<
t_{e,j}^{\mathrm{end}},\\
0,
& \text{otherwise}.
\end{cases}
\end{equation}
Here, \(t_{e,j}^{\mathrm{fail}}\) and \(t_{e,j}^{\mathrm{end}}\)
denote the failure occurrence time and failure end time stored in
\(F_{e,j}\). Thus, \(y_{e,t}=0\) denotes an operating instant,
including an instant within a degradation period, whereas \(y_{e,t}=1\)
denotes an instant within a failure interval.

The complete label sequence is
\begin{equation}
\label{eq:equipment-state-sequence}
\mathbf y^{(e)}
=
\left(
y_{e,1},
y_{e,2},
\ldots,
y_{e,T_e}
\right)^\top
\in\{0,1\}^{T_e}.
\end{equation}
At each \(t\in\{1,\ldots,T_e\}\), the released time-series row
contains the timestamp, the sensor-output vector \(X_t^{(e)}\), the
equipment-state label \(y_{e,t}\), and the annotation fields summarized
in Section~\ref{app:output-assembly} of the Appendix.

\subsection{Internal State and Calculation Rules}
\label{sec:internal-state}

The internal state \(Z_{e,t}\) contains the simulator variables required
for subsequent updates. It may include thermal, control, event, degradation,
history, cumulative-energy, humidity, and frost variables according to
the refrigerator configuration. Unlike the fixed configuration
\(\theta_e\), \(Z_{e,t}\) can change at each time instant and is not
included in the released files.

At each time instant, the equipment-state label is calculated first.
The internal state is then updated as
\begin{equation}
\label{eq:state-update-method}
Z_{e,t}
=
\begin{cases}
Z_{e,t-1},
& y_{e,t}=1,\\
u\!\left(
Z_{e,t-1},
\theta_e,
\mathcal F_e,
\mathcal D_e,
t,
\mathcal R_{e,t}
\right),
& y_{e,t}=0,
\end{cases}
\end{equation}
where \(u\) denotes the operating state-update function,
\(\mathcal D_e\) is the planned defrost schedule, and
\(\mathcal R_{e,t}\) groups the realizations of the random variables
required at time \(t\).

The sensor-output vector is generated from the current simulator state
according to
\begin{equation}
\label{eq:sensor-output-mapping-method}
X_t^{(e)}
=
\begin{cases}
g\!\left(
Z_{e,t},
\theta_e,
t,
\mathcal R_{e,t}
\right),
& y_{e,t}=0,\\
X_{t-1}^{(e)},
& y_{e,t}=1,
\end{cases}
\end{equation}
where \(g\) assembles the available sensor outputs. The fixed
configuration determines which optional outputs are present.

\subsection{Degradation and Failure Process}
\label{sec:degradation-failure-process}

During a degradation period, the refrigerator remains in operation while
the variables associated with the active degradation progressively affect
subsequent state updates and sensor outputs. At the first operating minute
after a failure interval, these variables return to their configured
initial values.

Six types of progressive degradation are included. Their selection,
timing, and associated simulator variables are described in
Section~\ref{app:failure-scheduling} of the Appendix.

\subsection{Generation Algorithm}
\label{sec:generation-algorithm}

Algorithm~\ref{alg:data_generation} summarizes the generation procedure
for one refrigerator.

\begin{algorithm}[H]
\caption{Generation of synthetic data for one refrigerator}
\label{alg:data_generation}
\begin{algorithmic}[1]

\Statex \textbf{Input} fixed refrigerator configuration \(\theta_e\)
and time-series length \(T_e\)

\Statex \textbf{Output} time-series table containing the timestamps,
\(\mathbf X^{(e)}\), and \(\mathbf y^{(e)}\), together with the
failure log \(\mathcal F_e\)

\State Construct \(\mathcal F_e\) using the scheduling procedure in
Section~\ref{app:failure-scheduling} of the Appendix

\State Construct the planned defrost schedule \(\mathcal D_e\) using
the defrost rule in Section~\ref{app:phenomena} of the Appendix

\State Initialize \(Z_{e,0}\) using the initial conditions defined with
the corresponding phenomena in Section~\ref{app:phenomena} of the Appendix

\For{\(t=1,\ldots,T_e\)}

    \State Calculate \(y_{e,t}\) using
    Eq.~\eqref{eq:equipment-state-label}

    \State Generate the realizations grouped in \(\mathcal R_{e,t}\)
    that are required by the active simulation rules

    \State Calculate \(Z_{e,t}\) using
    Eq.~\eqref{eq:state-update-method}

    \State Calculate \(X_t^{(e)}\) using
    Eq.~\eqref{eq:sensor-output-mapping-method}

\EndFor

\State Save \(\mathbf X^{(e)}\), \(\mathbf y^{(e)}\) and \(\mathcal F_e\)
\end{algorithmic}
\end{algorithm}

The computational cost depends on the operations performed at each
simulated minute. Let \(H_e^{\mathrm{hist}}\) denote the number of
past compressor-state values retained for the duty-cycle variable calculation. In the
current list-based implementation, the simulator may inspect the
\(M_e\) entries of the failure log \(\mathcal F_e\) and the
\(\lvert\mathcal D_e\rvert\) planned defrost starts at each time
instant. During an operating instant, it also processes up to
\(H_e^{\mathrm{hist}}\) compressor-history values and generates
\(m_e\) sensor outputs.

The complexity required to simulate one time instant is therefore bounded by
\[
\mathcal O\!\left(
M_e
+
\lvert\mathcal D_e\rvert
+
H_e^{\mathrm{hist}}
+
m_e
\right).
\]
Since this procedure is repeated for the \(T_e\) time instants of the
simulated record, the worst-case time complexity for one refrigerator is
\[
\mathcal O\!\left(
T_e
\left(
M_e
+
\lvert\mathcal D_e\rvert
+
H_e^{\mathrm{hist}}
+
m_e
\right)
\right).
\]

\subsection{Illustrative Example}
\label{sec:illustrative-example}

Consider a refrigerator with a fridge compartment and thermostat
thresholds
\[
\tau_{e,\mathrm{fr}}^{\mathrm{low}}=3.5\,^{\circ}\mathrm{C},
\qquad
\tau_{e,\mathrm{fr}}^{\mathrm{high}}=5.0\,^{\circ}\mathrm{C}.
\]
Suppose that its failure log contains a retained failure associated with
\texttt{COMP\_DEGRAD} degradation, with
\[
t_{e,1}^{\mathrm{deg}}=43{,}201,
\qquad
t_{e,1}^{\mathrm{fail}}=50{,}401,
\qquad
t_{e,1}^{\mathrm{end}}=51{,}001.
\]

At \(t=100\), this failure is not active and the refrigerator is
operating, so \(y_{e,100}=0\). If the fridge temperature at the
previous instant is
\[
T_{e,99}^{\mathrm{fr}}=5.4\,^{\circ}\mathrm{C},
\]
the upper thermostat threshold is exceeded and the compressor is
activated. Under nominal compressor and fan efficiencies and with the
temperature perturbation set to zero for this illustration, the
fridge-temperature rule in
Section~\ref{app:thermal-behaviour} of the Appendix gives
\[
T_{e,100}^{\mathrm{fr}}
=
5.4-0.25
=
5.15\,^{\circ}\mathrm{C}.
\]
The updated fridge temperature and compressor state are stored in
\(Z_{e,100}\) and emitted as the corresponding entries of
\(X_{100}^{(e)}\).
At \(t=50{,}700\),
\[
t_{e,1}^{\mathrm{fail}}
\leq
t
<
t_{e,1}^{\mathrm{end}},
\]
so \(y_{e,50{,}700}=1\). Equations~\eqref{eq:state-update-method}
and~\eqref{eq:sensor-output-mapping-method} therefore retain
\(Z_{e,50{,}699}\) and \(X_{50{,}699}^{(e)}\) at this instant.

\section{Technical Validation}
\label{sec:technical_validation}

Technical validation uses a selected one-year time series containing
525{,}600 minute-level rows together with dataset-level analyses of all
27 time-series files and failure logs.

\subsection{Nominal Operating Behaviour}
\label{sec:validation-baseline}

\begin{figure}[!htbp]
    \centering
    \includegraphics[width=\linewidth]
    {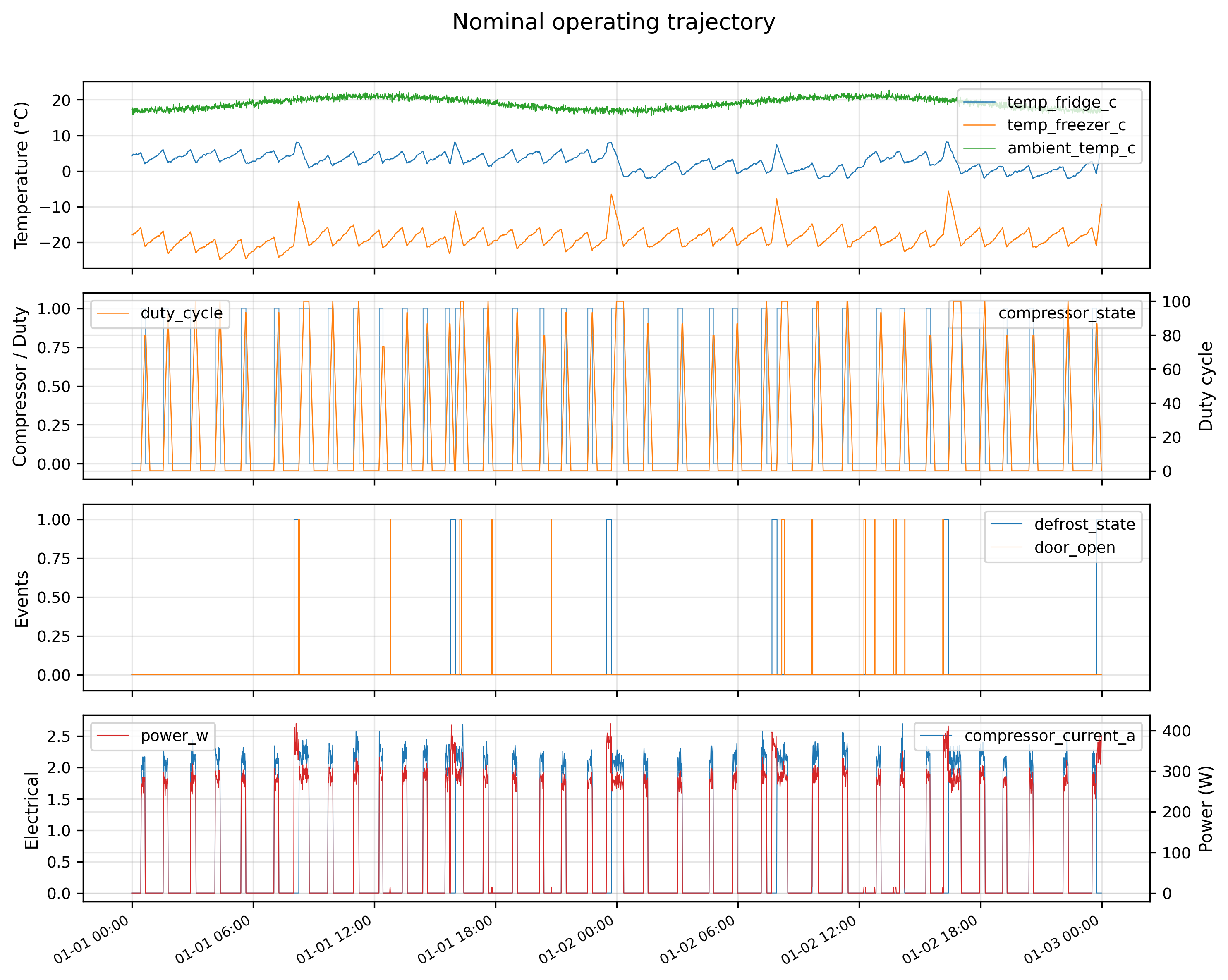}
    \caption{Selected 48-hour nominal operating period. The panels show
    compartment temperatures, compressor and defrost operation,
    door-opening events, and electrical load.}
    \label{fig:nominal}
\end{figure}

The interactions between the main simulated phenomena were examined
over a 48-hour window randomly selected from periods without degradation
or failure. Figure~\ref{fig:nominal} shows the
compartment temperatures, compressor and defrost states, door openings,
and electrical load over the selected period. The compressor is active
during 21.0\% of the period, which contains 13 door-opening events and
6 defrost events. Compartment temperatures cycle with compressor
switching, defrost interrupts compressor operation, and door-opening
times can be compared with the temperature and electrical outputs.

\subsection{Pre-Failure Behaviour}
\label{sec:validation-prefailure}

\begin{figure}[!htbp]
    \centering
    \includegraphics[width=\linewidth]
    {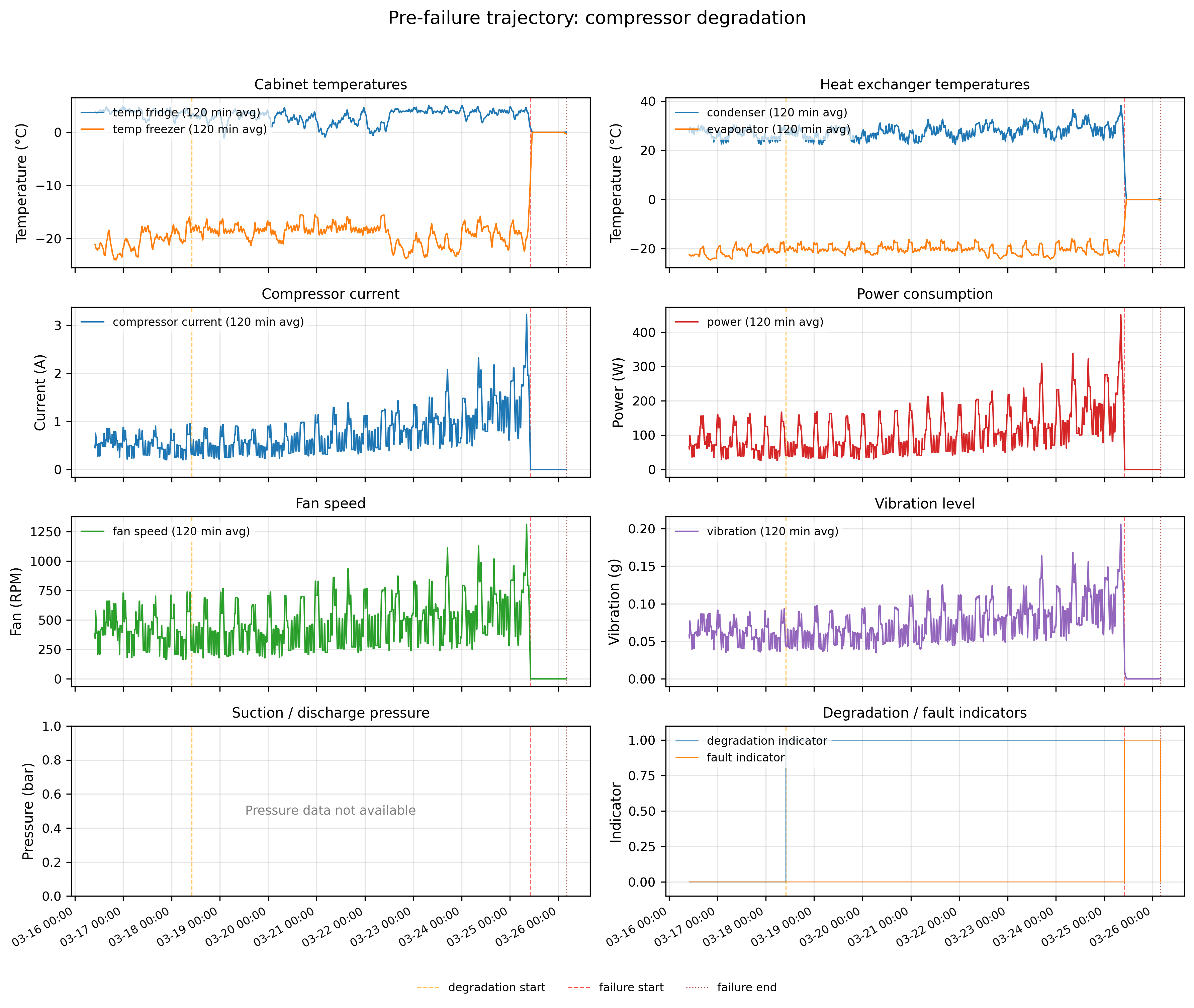}
    \caption{Sensor outputs around a compressor-efficiency failure. The
    vertical markers identify degradation start, failure occurrence, and
    failure end. The pre-failure analysis uses the portion ending at
    failure occurrence. The values are two-hour moving averages.}
    \label{fig:compdegrad}
\end{figure}

The effect of degradation on the generated sensor outputs before failure
occurrence was examined using one configured compressor-efficiency
degradation, \cfgfield{COMP_DEGRAD} in
Table~\ref{tab:degradation-types}. Figure~\ref{fig:compdegrad} shows
the period from two days
before degradation start to failure occurrence. The displayed time
series were smoothed with a two-hour moving window to reduce rapid
fluctuations and make the pre-failure trends easier to observe. After degradation start, the two-hour moving average of compressor
current increases from 0.81\,A to 2.3\,A near failure occurrence,
while local maxima in the displayed curve approach 3\,A. Over the same interval, the
fridge-compartment temperature decreases from
3.77 to 3.01\,$^\circ$C. The other panels show the outputs generated over the same interval.
No additional numerical trend estimates are reported for these outputs. Under the implemented rules, reduced compressor efficiency lowers the
cooling rate and thermostat control can keep the compressor active for
longer. The failure occurrence time is configured in advance and is
not triggered by the sensor outputs.

\subsection{Class Composition and Failure Durations}
\label{sec:validation-composition}

\begin{figure}[!htbp]
    \centering
    \includegraphics[width=\linewidth]
    {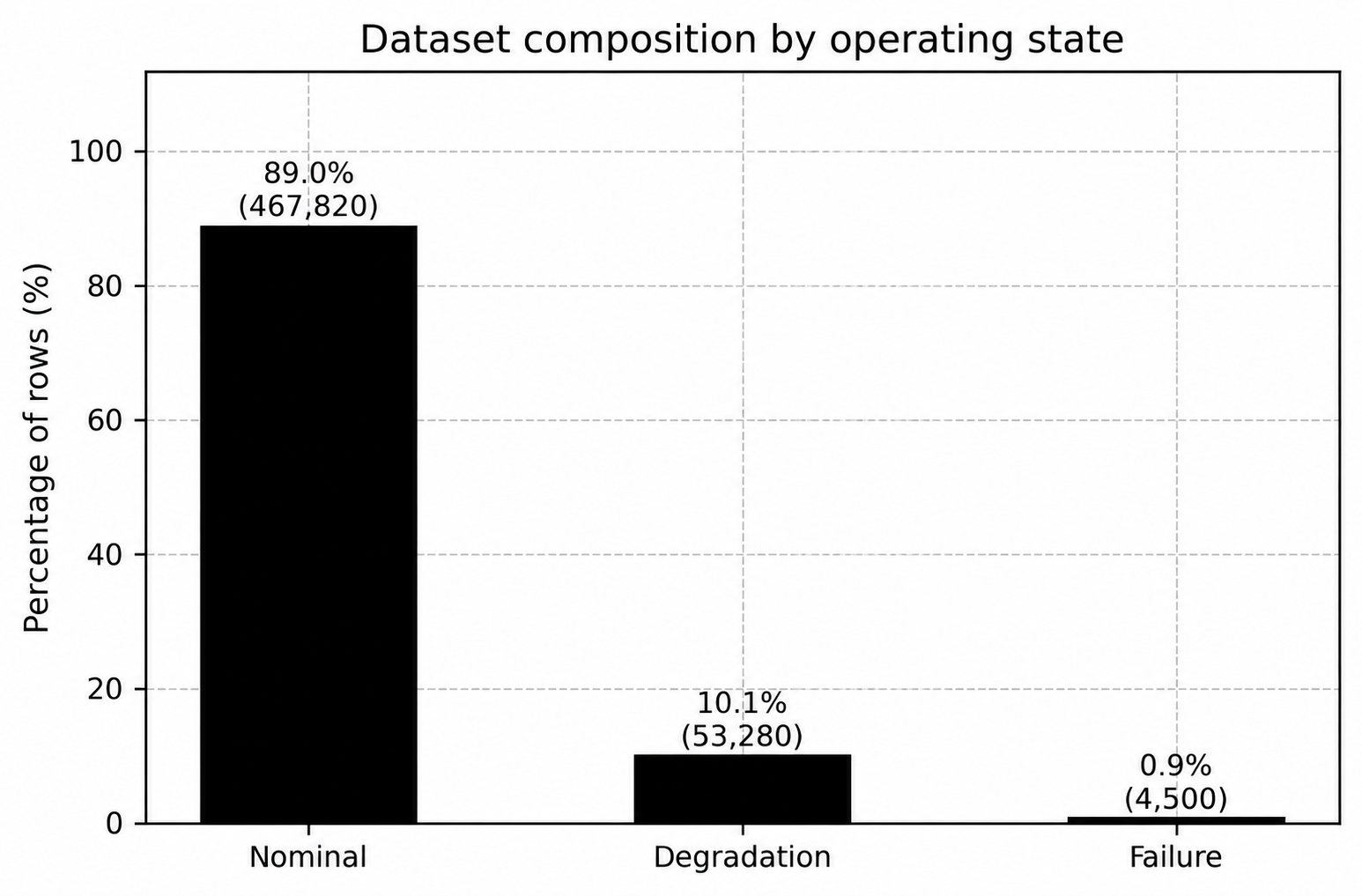}
    \caption{Class composition of the selected time series. The
    percentages refer to observations assigned to nominal
    operation, degradation, or failure.}
    \label{fig:composition}
\end{figure}

The proportions of nominal, degradation, and failure observations were
quantified together with the durations assigned to the generated
failures. Figure~\ref{fig:composition} gives the class composition of
the selected time series. Nominal operation accounts for 89.0\% of its
minute-level rows, degradation for 10.1\%, and failure for 0.9\%.
The series contains 7 failure occurrences, with a mean degradation
duration of 5.3 days and a mean failure-interval duration of 10.7 hours.

\begin{figure}[!htbp]
    \centering
    \includegraphics[width=\linewidth]
    {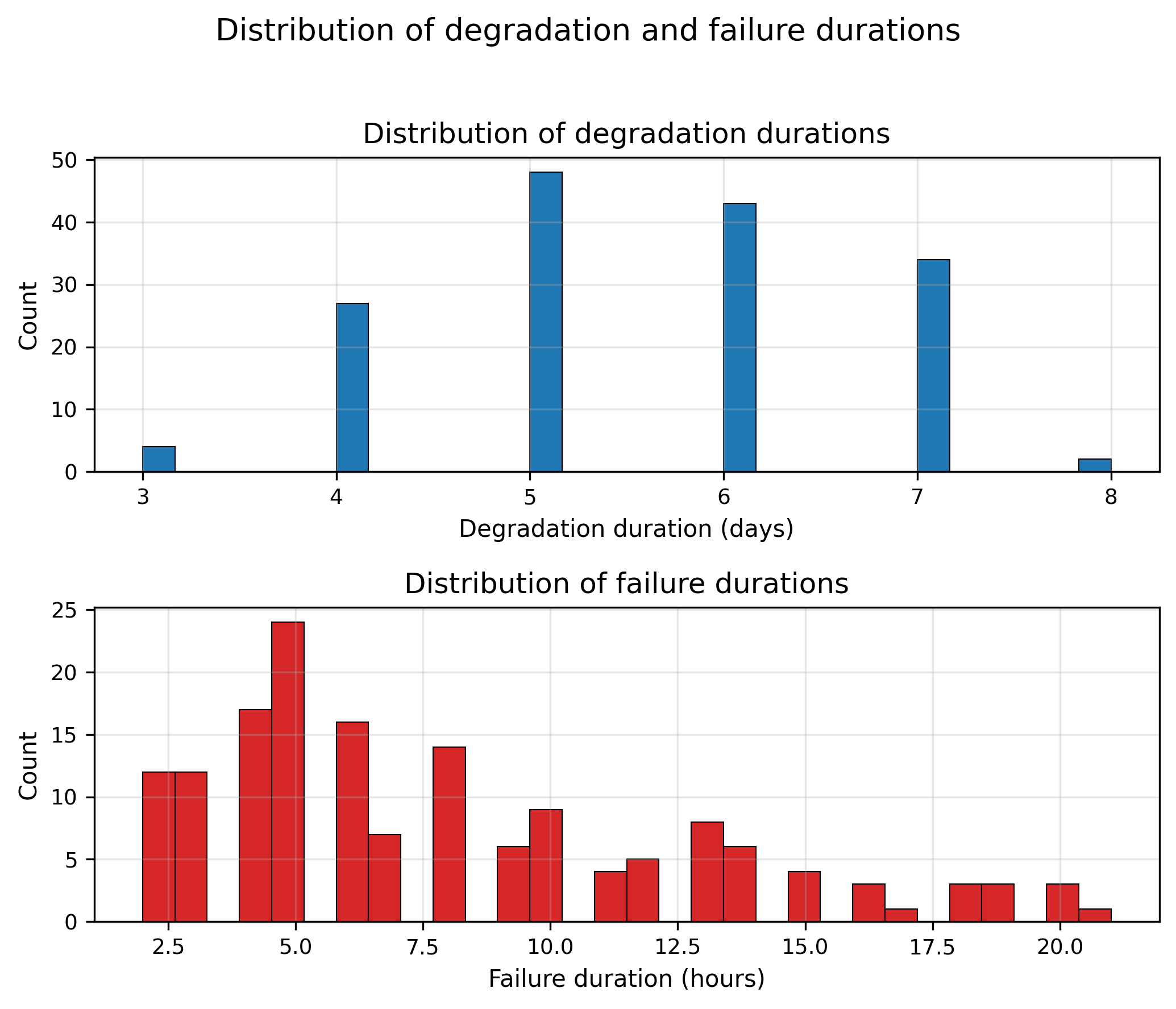}
    \caption{Distributions of degradation and failure-interval durations across
    the 27 refrigerators.}
    \label{fig:durations}
\end{figure}

Across the 27 refrigerators, the dataset contains 158 failure
occurrences. Each occurrence has one degradation period and one failure interval. Figure~\ref{fig:durations} shows the two distributions.
Degradation durations range from 3.0 to 8.0 days, with a mean of
5.52 days and a median of 5.50 days. Failure-interval durations range from
2.0 to 21.0 hours. These durations correspond to the simulation
settings used to generate the dataset and are not presented as measured
real-world failure timescales.

\subsection{Sensor-Output Availability}
\label{sec:validation-heterogeneity}

\begin{figure}[!htbp]
    \centering
    \includegraphics[width=0.8\linewidth]
    {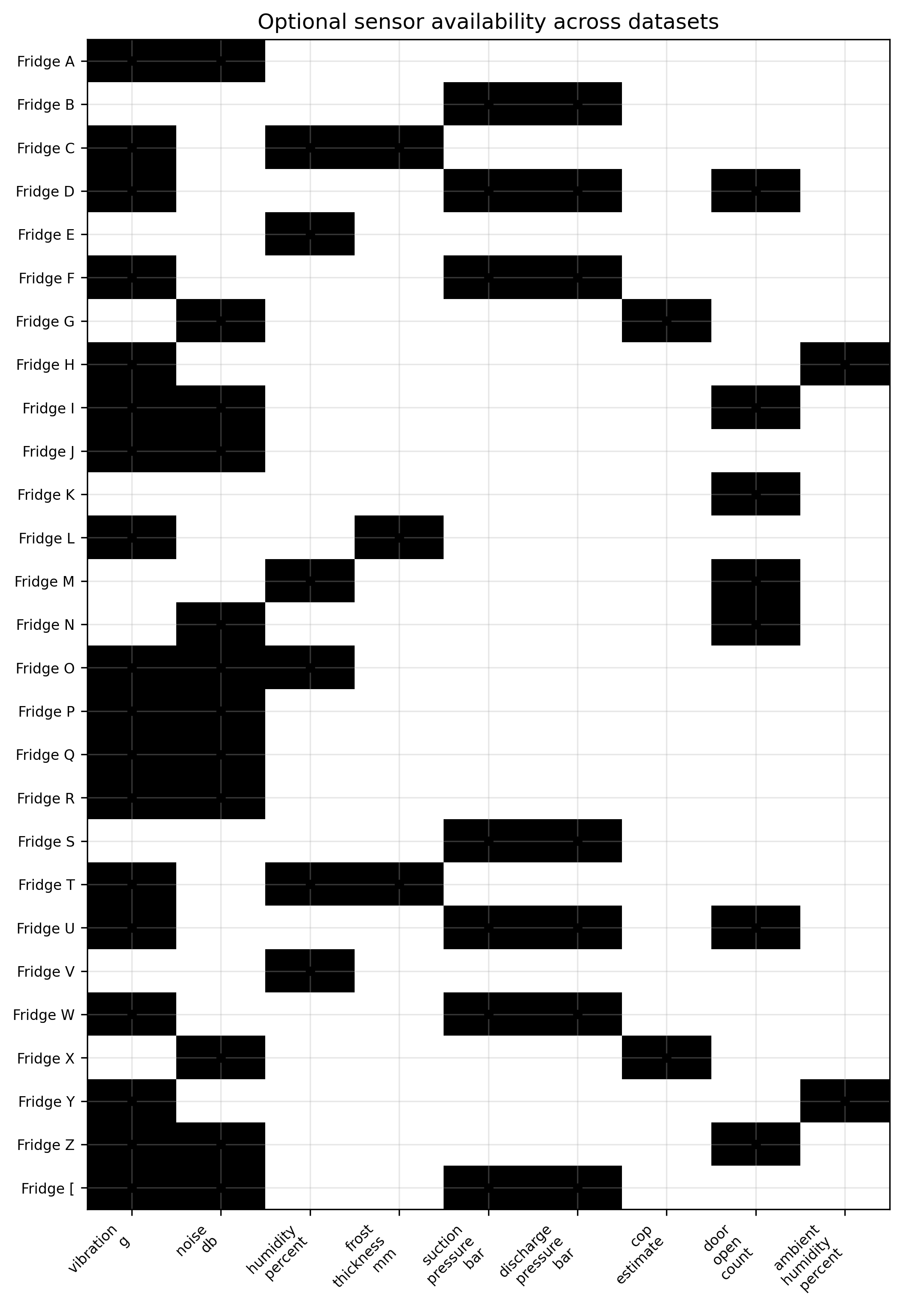}
    \caption{Availability of optional sensor outputs across the
    27 refrigerators.}
    \label{fig:heterogeneity}
\end{figure}

Sensor-output availability was examined to quantify the differences in
optional sensor outputs across the simulated refrigerators. For an
optional sensor output, availability is defined as the proportion of
the 27 refrigerators whose released time-series table contains the
corresponding output column. Figure~\ref{fig:heterogeneity} shows these
proportions across the equipment family.

The most frequent optional sensor output is
\cfgfield{vibration_g}, which is present for 66.7\% of the
refrigerators, followed by \cfgfield{noise_db}, which is present for
44.4\%. The outputs \cfgfield{suction_pressure_bar} and
\cfgfield{discharge_pressure_bar} are each present for 25.9\% of the
refrigerators. The corresponding proportions are 22.2\% for
\cfgfield{humidity_percent}, 11.1\% for
\cfgfield{frost_thickness_mm}, and 7.4\% for both
\cfgfield{cop_estimate} and
\cfgfield{ambient_humidity_percent}. These differences produce
refrigerators with different sensor-output sets.

\subsection{Failure-Prediction Evaluation}
\label{sec:failure-prediction-evaluation}

The generated time series were evaluated to determine whether their
pre-failure sensor-output changes can be used by a prediction model to
anticipate future failures. The evaluation uses the transferable autologistic
failure-prediction model. The common model is estimated
from 17 training refrigerators and then applied directly to the target
refrigerators. For each target refrigerator, the target-specific
parameters are re-estimated using the common-model parameters as prior
information, following the adaptation procedure described
in~\cite{benamirouche2026transferable}. The resulting target-specific
model is evaluated on the corresponding target refrigerator. The
evaluation includes 10 target refrigerators containing 59 failure
occurrences.

Table~\ref{tab:contextual-prediction} summarizes the results obtained
with the two model settings.

\begin{table}[htbp]
\centering
\caption{Failure-prediction results on the 10 target refrigerators.}
\label{tab:contextual-prediction}
\begin{tabular}{lcc}
\toprule
Metric & Common model & Target-specific model \\
\midrule
Detected failures & 36/59 & 54/59 \\
Missed failures & 23/59 & 5/59 \\
Detection rate (\%) & 61.0 & 91.5 \\
False alerts & 24 & 13 \\
Mean lead time (h) & 132.8 & 101.2 \\
\bottomrule
\end{tabular}
\end{table}

Target-specific adaptation increased the detection rate while reducing
missed failures and false alerts, indicating in this evaluation that the
generated pre-failure sensor-output changes can be used for prediction.

\section{Conclusion}
\label{sec:conclusion}

In this work, we present a physics-inspired failure simulator for 27
refrigerators that addresses the limited availability of datasets
describing refrigerator faults and their causes. The simulator
generates synthetic data based on the physical degradation mechanisms
of components within the refrigerators, while providing controlled
degradation evolution and failure timing across heterogeneous sensor
configurations. The resulting dataset provides a reproducible basis
for developing and evaluating methods for fault diagnosis, failure
prediction, and prognostics, particularly in contexts where detailed
real-world fault data are scarce. By linking sensor observations to
the physical mechanisms underlying component degradation, the proposed
approach supports the development of data-driven methods for detecting
degradation, anticipating failures, and predicting their occurrence.
Future work will extend the physical interactions represented in the
simulator and compare selected simulated phenomena with experimental
and operational measurements. Further validation against real
degradation and failure histories will depend on the availability of
sufficiently detailed and well-documented refrigerator datasets.

\section*{Data Availability}

The complete dataset is publicly available on Zenodo
\cite{BenamiroucheDataset2026}. The deposited data include the
27 refrigerator time-series files and their corresponding failure logs.

\section*{Code Availability}

The Python generation code, refrigerator configurations, and
documentation are publicly available at
\url{https://github.com/iben0x/fridge-sensor-data}.

\section*{Ethics Statement}

The dataset is entirely synthetic and contains no personal data, no patient data, and no human-subject information.

\clearpage

\appendix

\section*{Appendix}
\addcontentsline{toc}{section}{Appendix}

This appendix complements the data-generation method described in
Section~\ref{sec:methods} by providing the definitions and equations
needed to reproduce the simulator. It describes the refrigerator
components included in the simulator, the failure and degradation
scheduling procedure, the simulation rules, and the assembly of the
generated outputs. The notation introduced in
Section~\ref{sec:methods} is retained, while additional symbols are
defined at their first use. The superscripts
\(\mathrm{deg}\), \(\mathrm{fail}\), and \(\mathrm{end}\) refer to
degradation start, failure occurrence, and failure end. The
abbreviations \(\mathrm{fr}\), \(\mathrm{fz}\), \(\mathrm{ev}\),
\(\mathrm{cd}\), and \(\mathrm{amb}\) refer to the fridge compartment,
freezer compartment, evaporator, condenser, and ambient conditions,
respectively. Similarly, \(\mathrm{comp}\), \(\mathrm{def}\),
\(\mathrm{door}\), \(\mathrm{mec}\), and \(\mathrm{th}\) refer to
compressor, defrost, door, mechanical, and thermal quantities.
Finally, \(\mathcal N(\mu,\sigma^2)\) denotes a Gaussian distribution
with mean \(\mu\) and variance \(\sigma^2\),
\(\operatorname{Bernoulli}(p)\) denotes a Bernoulli distribution with
parameter \(p\), and
\(\mathcal U_{\mathrm{disc}}(a,b)\) denotes the discrete uniform
distribution over the integers from \(a\) to \(b\), inclusive.

\section{Refrigerator Components and Their Roles}
\label{app:refrigerator-interactions}

\begin{figure}[H]
    \centering
    \includegraphics[width=0.6\textwidth]
    {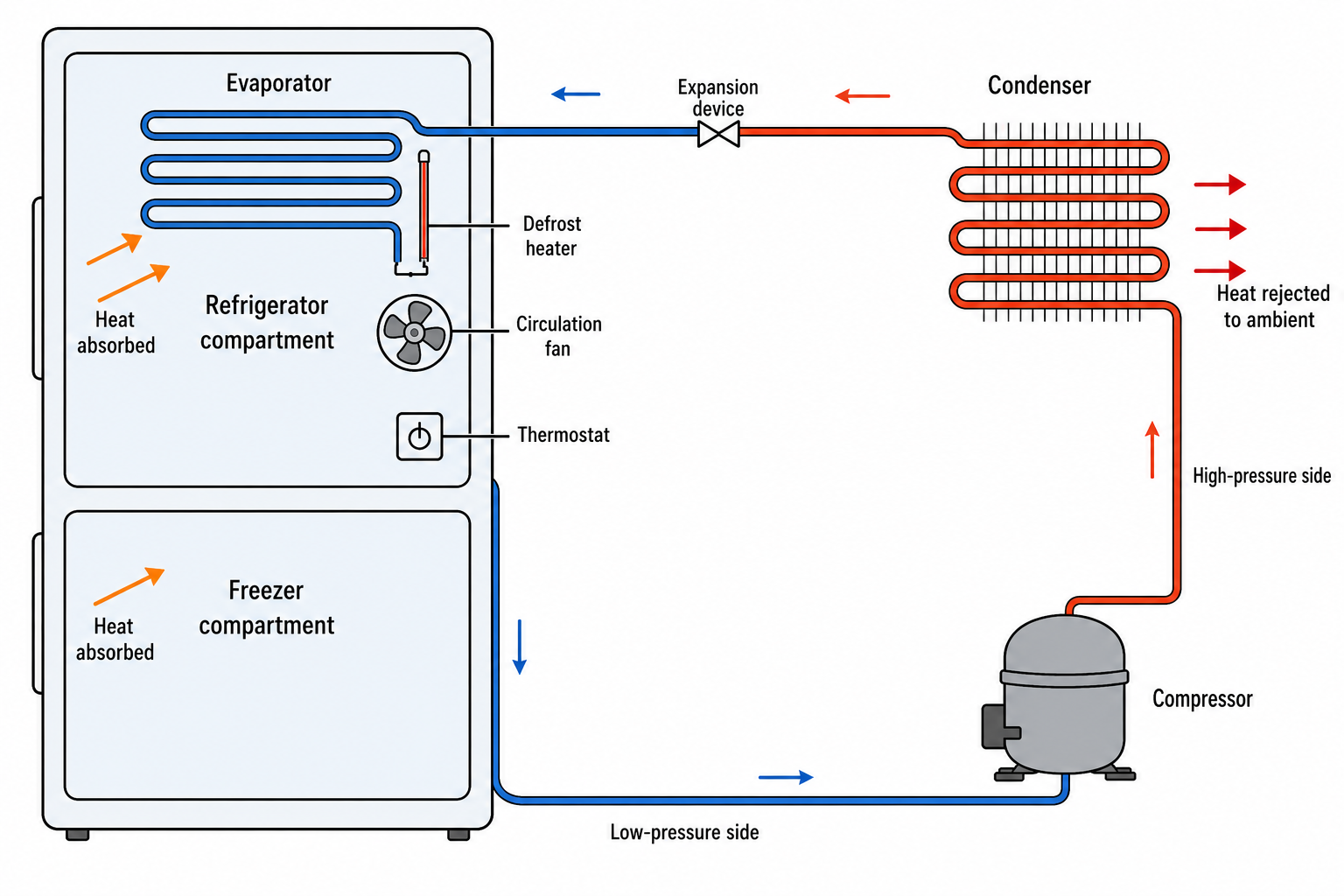}
    \caption{Simplified schematic of a two-compartment household
    refrigerator and its vapour-compression circuit.}
    \label{fig:refrigerator-cycle}
\end{figure}

A household refrigerator transfers heat from its refrigerated
compartments to the ambient environment through a closed
vapour-compression refrigeration circuit. A refrigerant circulates
continuously through this circuit and changes pressure, temperature,
and physical state between liquid and vapour as it passes through the
compressor, condenser, expansion device, and evaporator. The compressor
draws low-pressure refrigerant vapour from the evaporator and increases
its pressure and temperature. The condenser then rejects heat to the
ambient environment and condenses the refrigerant into a liquid. The
expansion device reduces the refrigerant pressure before it enters the
evaporator, where heat is absorbed from the refrigerated compartments
and the refrigerant evaporates before returning to the compressor
\cite{ASHRAE2022}. The main components and the direction of refrigerant
flow are shown in Figure~\ref{fig:refrigerator-cycle}. The low- and
high-pressure sides of the refrigerant circuit are shown in blue and
red, respectively.

These components affect refrigerator operation in different ways. The refrigerated cabinet limits heat transfer from the ambient
environment, while the door provides access to the compartments and
the door seal limits the entry of warm and moist air when the door is
closed \cite{LiuGasket2021}. Opening the door or degrading the seal can
therefore increase compartment temperature, cooling demand, internal
humidity, and frost formation. Temperature is regulated through the
thermostat and compressor. The thermostat compares compartment
temperature with configured thresholds and switches the compressor on
or off accordingly \cite{Caglayan2022}. When the compressor is active,
the evaporator becomes cold and removes heat from the refrigerated
compartments; when it is inactive, compartment and evaporator
temperatures gradually move toward warmer surrounding temperatures.
Fans, when present, circulate air across the heat exchangers and
through the refrigerated compartments, supporting heat transfer and
temperature distribution \cite{Yoo2024}. A reduction in fan
performance can therefore reduce cooling effectiveness and affect fan
speed, vibration, acoustic noise, and refrigerator energy performance.

Frost can accumulate on the evaporator and reduce airflow and heat
transfer. Frost-free refrigerators therefore use a defrost heater to
periodically warm the evaporator and remove accumulated frost
\cite{ZhangDefrost2025}. During a defrost cycle, normal cooling is
temporarily modified, while reduced defrost-heater performance can
decrease the effectiveness of frost removal. The compressor,
condenser, fan, door seal, and defrost heater are also the components
affected by the progressive degradations generated in the dataset.
Changes in their condition can influence several quantities at the
same time. For example, compressor degradation can affect cooling,
current, vibration, acoustic noise, pressure, and estimated COP, while
condenser degradation can increase condenser temperature and discharge
pressure. These interactions provide the physical motivation for the
simulator rules presented in Section~\ref{app:phenomena} of the
Appendix. 
\section{Failure and Degradation Scheduling}
\label{app:failure-scheduling}

For each refrigerator, the simulator determines how many failures
occur, their degradation types and durations, and when degradation and
failure occur. The corresponding information is stored in the failure
log \(\mathcal F_e\). The scheduling parameters are components of
\(\theta_e\). These rules define the timing of the failures, while
Section~\ref{app:phenomena} of the Appendix describes their effects on refrigerator
operation.

\subsection{Failure Selection and Duration}
\label{app:failure-selection-duration}

\paragraph{Number of failures:}
The configuration specifies the minimum and maximum numbers of
failures, \(n_e^{\min}\) and \(n_e^{\max}\). The requested number of
failures is drawn uniformly within this range,
\begin{equation}
N_e^{\mathrm{req}}
\sim
\mathcal U_{\mathrm{disc}}
\left(
n_e^{\min},
n_e^{\max}
\right).
\end{equation}
The simulator attempts to schedule \(N_e^{\mathrm{req}}\) failures,
although fewer may be retained if there is not enough simulation time.

\paragraph{Degradation type:}
Each failure is associated with one of six degradation types. Each
type modifies one or more internal variables during the degradation
period. These variables are the compressor-efficiency factor
\(\eta_{e,t}^{\mathrm{comp}}\), the compressor mechanical-wear level
\(w_{e,t}^{\mathrm{mec}}\), the condenser thermal-resistance factor
\(R_{e,t}^{\mathrm{th}}\), the door-seal efficiency
\(\eta_{e,t}^{\mathrm{seal}}\), the fan efficiency
\(\eta_{e,t}^{\mathrm{fan}}\), and the defrost-heater efficiency
\(\eta_{e,t}^{\mathrm{def}}\). Table~\ref{tab:degradation-types}
shows the variables modified by each degradation type and their main
effects on the generated outputs.

\begin{table}[H]
\centering
\small
\caption{Progressive degradation types included in the generated data.}
\label{tab:degradation-types}
\begin{tabular}{
>{\raggedright\arraybackslash}p{0.22\linewidth}
>{\raggedright\arraybackslash}p{0.16\linewidth}
>{\raggedright\arraybackslash}p{0.18\linewidth}
>{\raggedright\arraybackslash}p{0.34\linewidth}}
\toprule
Degradation type &
Code &
Modified variables &
Main effects on generated outputs \\
\midrule

Compressor efficiency loss &
\cfgfield{COMP_DEGRAD} &
\(\eta_{e,t}^{\mathrm{comp}}\) &
Reduced cooling effectiveness; increased compressor current,
condenser temperature, suction and discharge pressure, vibration,
acoustic noise, and electrical power; reduced estimated COP. \\

Compressor mechanical wear &
\cfgfield{COMP_WEAR} &
\(w_{e,t}^{\mathrm{mec}}\),
\(\eta_{e,t}^{\mathrm{comp}}\) &
Increased compressor current, vibration, and acoustic noise, with
reduced cooling effectiveness and estimated COP. \\

Condenser fouling &
\cfgfield{COND_DIRT} &
\(R_{e,t}^{\mathrm{th}}\) &
Increased condenser temperature, compressor current, discharge
pressure, electrical power, and acoustic noise; reduced estimated COP. \\

Door-seal degradation &
\cfgfield{DOOR_SEAL} &
\(\eta_{e,t}^{\mathrm{seal}}\) &
Increased compartment warming, internal humidity, frost accumulation,
and simulated opening events associated with seal degradation. \\

Fan degradation &
\cfgfield{FAN_DEGRAD} &
\(\eta_{e,t}^{\mathrm{fan}}\) &
Reduced fan speed and fridge cooling rate, with changes in vibration
and acoustic noise and reduced estimated COP. \\

Defrost-heater degradation &
\cfgfield{DEFROST_FAIL} &
\(\eta_{e,t}^{\mathrm{def}}\) &
Reduced defrost-heater current and frost removal, with increased frost
accumulation. \\

\bottomrule
\end{tabular}
\end{table}

Effects involving optional sensor outputs apply only when the
corresponding outputs are enabled in \(\theta_e\).

A refrigerator does not need to include all six degradation types.
Let \(K_e\) be the number of available types and
\(\omega_{e,r}\) the selection weight of type \(r\). For failure \(j\),
the probability of selecting type \(r\) is
\begin{equation}
\Pr\!\left(
\text{type }r\text{ is selected for failure }j
\right)
=
\frac{\omega_{e,r}}
{\displaystyle\sum_{r'=1}^{K_e}\omega_{e,r'}}.
\end{equation}
A larger weight makes the corresponding degradation type more likely
to be selected.

\paragraph{Durations:}
After selecting the degradation type, the simulator generates the
degradation duration and the failure-interval duration. The
degradation duration is drawn as an integer number of days,
\begin{equation}
D_{e,j}^{\mathrm{deg}}
\sim
\mathcal U_{\mathrm{disc}}
\left(
d_{e,r}^{\min},
d_{e,r}^{\max}
\right),
\end{equation}
while the failure duration is drawn as an integer number of hours,
\begin{equation}
L_{e,j}^{\mathrm{fail}}
\sim
\mathcal U_{\mathrm{disc}}
\left(
\ell_{e,r}^{\min},
\ell_{e,r}^{\max}
\right).
\end{equation}
The two durations are then converted to minutes,
\begin{equation}
\delta_{e,j}^{\mathrm{deg}}
=
1440D_{e,j}^{\mathrm{deg}},
\qquad
\ell_{e,j}^{\mathrm{fail}}
=
60L_{e,j}^{\mathrm{fail}}.
\end{equation}

\subsection{Failure Placement}
\label{app:failure-placement-retention}

\paragraph{Degradation start:}
Let \(a_{e,j}\) denote the reference day from which failure \(j\) is
scheduled. For the first failure,
\begin{equation}
a_{e,1}
=
d_e^{\mathrm{ref}}.
\end{equation}

The actual degradation start is shifted from this reference day by an
additional number of days,
\begin{equation}
G_{e,j}
\sim
\mathcal U_{\mathrm{disc}}
\left(
G_e^{\min},
G_e^{\min}+20
\right).
\end{equation}
The 20-day range introduces variability in the spacing between
degradation periods while preserving the configured minimum gap.

An hour is also drawn uniformly so that degradation starts are
distributed throughout the day rather than occurring at a fixed hour,
\begin{equation}
H_{e,j}
\sim
\mathcal U_{\mathrm{disc}}(0,23).
\end{equation}

\paragraph{Available simulation time:}
Before retaining a failure, the simulator checks that both its
degradation period and failure interval fit within the available
record. The last day available for scheduling is
\begin{equation}
d_e^{\mathrm{hor}}
=
\min\!\left(
\frac{T_e}{1440},
d_e^{\max}
\right).
\end{equation}
The failure is retained only if
\begin{equation}
a_{e,j}
+
G_{e,j}
+
\frac{H_{e,j}}{24}
+
D_{e,j}^{\mathrm{deg}}
+
\frac{L_{e,j}^{\mathrm{fail}}}{24}
\le
d_e^{\mathrm{hor}}.
\label{eq:failure-admissibility}
\end{equation}
This condition ensures that the complete degradation and failure
interval remains within the scheduling horizon. If it is not
satisfied, scheduling stops. The final number of failures can therefore
be smaller than the number initially requested.

\paragraph{Failure times:}
For each retained failure, the degradation start is
\begin{equation}
t_{e,j}^{\mathrm{deg}}
=
1
+
1440\left(a_{e,j}+G_{e,j}\right)
+
60H_{e,j}.
\end{equation}
The additional 1 accounts for the one-based time indexing used in the
generated time series.

The failure occurrence and failure end are then
\begin{align}
t_{e,j}^{\mathrm{fail}}
&=
t_{e,j}^{\mathrm{deg}}
+
\delta_{e,j}^{\mathrm{deg}},
\\
t_{e,j}^{\mathrm{end}}
&=
t_{e,j}^{\mathrm{fail}}
+
\ell_{e,j}^{\mathrm{fail}}.
\end{align}

\paragraph{Next failure:}
After a retained failure, the simulator adds a two-day separation
before scheduling the next one. This prevents successive failure
intervals from being placed immediately next to each other,
\begin{equation}
a_{e,j+1}
=
a_{e,j}
+
G_{e,j}
+
\frac{H_{e,j}}{24}
+
D_{e,j}^{\mathrm{deg}}
+
\frac{L_{e,j}^{\mathrm{fail}}}{24}
+
2.
\end{equation}
New values of \(G_{e,j+1}\) and \(H_{e,j+1}\) are then generated for
the next failure.

\subsection{Degradation Progress and Failure Log}
\label{app:degradation-progress-log}

The number of retained failures is denoted by \(M_e\). Because some
requested failures may not fit within the simulation period,
\begin{equation}
M_e
\le
N_e^{\mathrm{req}}.
\end{equation}

\paragraph{Degradation progress:}
During a degradation period, \(P_{e,j,t}\) indicates how far the
degradation has progressed. It starts at 0 at degradation start and
reaches 1 at failure occurrence,
\begin{equation}
P_{e,j,t}
=
\frac{
t-t_{e,j}^{\mathrm{deg}}
}{
t_{e,j}^{\mathrm{fail}}
-
t_{e,j}^{\mathrm{deg}}
},
\qquad
t_{e,j}^{\mathrm{deg}}
\le
t
\le
t_{e,j}^{\mathrm{fail}}.
\label{eq:degradation-progress-appendix}
\end{equation}
This value is used by the degradation equations in
Section~\ref{app:phenomena} of the Appendix to progressively modify the corresponding
internal variables.

\paragraph{Failure log:}
Each retained failure produces one entry \(F_{e,j}\) in the failure
log \(\mathcal F_e\). Table~\ref{tab:failure-log-fields} lists the
stored fields.

\begin{table}[H]
\centering
\small
\caption{Fields included in each retained failure-log entry.}
\label{tab:failure-log-fields}
\begin{tabular}{p{0.27\linewidth} p{0.24\linewidth} p{0.39\linewidth}}
\toprule
Exported field & Value type and unit & Meaning \\
\midrule

\cfgfield{fault_id} & integer &
Failure identifier. \\

\cfgfield{fault_type} & text &
Descriptive failure type. \\

\cfgfield{fault_code} & categorical &
Degradation code listed in Table~\ref{tab:degradation-types}. \\

\cfgfield{severity} & categorical &
\texttt{Mineure}, \texttt{Majeure}, or \texttt{Critique}. \\

\cfgfield{degradation_start} & timestamp &
Degradation start time. \\

\cfgfield{degradation_days} & integer, days &
Degradation duration. \\

\cfgfield{failure_start} & timestamp &
Failure occurrence time. \\

\cfgfield{failure_end} & timestamp &
Failure end time. \\

\cfgfield{failure_duration_hours} & integer, hours &
Failure-interval duration. \\

\cfgfield{description} & text &
Short description of the failure. \\

\bottomrule
\end{tabular}
\end{table}

\section{Simulation Rules for Included Phenomena}
\label{app:phenomena}

The following subsections explain how the simulator calculates each
phenomenon and sensor output. Together, these rules define the
state-update function \(u\) in Eq.~\eqref{eq:state-update-method}.

\subsection{Ambient Conditions}
\label{app:ambient-conditions}

The ambient conditions describe the air surrounding the refrigerator.
Air temperature generally follows seasonal and daily patterns. It is
typically warmer in summer than in winter and warmer during the day
than at night \cite{Matthew2022,North2021}. Relative humidity also
changes over time and often follows an opposite daily pattern, with
higher values during cooler periods and lower values during warmer
periods \cite{Matthew2022,Shih2025}. The simulator uses these patterns
to generate ambient temperature and, when enabled in \(\theta_e\),
ambient relative humidity.

Because the simulation uses a one-minute time step, each time instant
\(t\) is converted into elapsed days \(d_t\) and the hour \(h_t\)
within the current day,
\begin{equation}
d_t=\frac{t-1}{1440},
\qquad
h_t=\frac{(t-1)\bmod 1440}{60},
\qquad
h_t\in[0,24).
\end{equation}
The value \(d_t\) is used for yearly variations, while \(h_t\) is used
for variations within a day.

\paragraph{Ambient temperature:}
Periodic functions have been used to describe repeated seasonal and
daily temperature variations \cite{North2021,Shah2024}. The simulator
therefore combines a reference temperature, a seasonal variation, a
daily variation, and a small random variation.

For refrigerator \(e\), the reference ambient temperature is
\(T_{e,\mathrm{base}}^{\mathrm{amb}}\). The seasonal amplitude
\(A_e^{\mathrm{seas}}\) controls the temperature
change over the year, while the daily amplitude
\(A_e^{\mathrm{day}}\) controls the change within a
day. These quantities are expressed in \(^{\circ}\mathrm{C}\).

A small random variation  $\varepsilon_{e,t}^{\mathrm{amb}}$, expressed in
\(^{\circ}\mathrm{C}\), is added to avoid perfectly repeated
temperature cycles,
\begin{equation}
\varepsilon_{e,t}^{\mathrm{amb}}
\sim
\mathcal N\!\left(
0,
(\sigma_e^{\mathrm{amb}})^2
\right),
\end{equation}
where \(\sigma_e^{\mathrm{amb}}\) controls the minute-to-minute
variation.

The ambient temperature is then calculated as
\begin{equation}
T_{e,t}^{\mathrm{amb}}
=
T_{e,\mathrm{base}}^{\mathrm{amb}}
+
A_e^{\mathrm{seas}}
\sin\!\left(
\frac{2\pi d_t}{365}
-
\frac{\pi}{2}
\right)
+
A_e^{\mathrm{day}}
\sin\!\left(
\frac{2\pi(h_t-6)}{24}
\right)
+
\varepsilon_{e,t}^{\mathrm{amb}}.
\end{equation}

The first periodic term produces the yearly temperature variation,
while the second produces the daily variation.

\paragraph{Ambient humidity:}
The simulator uses a reference ambient humidity
\(H_{e,\mathrm{base}}^{\mathrm{amb}}\) and adds also yearly, daily, and
random variations. To keep the model simple, two reference humidity
values are used according to the reference ambient temperature \(T_{e,\mathrm{base}}^{\mathrm{amb}}\),
\begin{equation}
H_{e,\mathrm{base}}^{\mathrm{amb}}
=
\begin{cases}
75,
& T_{e,\mathrm{base}}^{\mathrm{amb}}>28,\\
55,
& \text{otherwise}.
\end{cases}
\end{equation}
The values are expressed as percentages. This equation provides a
higher reference humidity for warmer ambient conditions and a lower
one for more moderate conditions.

A small random variation $\varepsilon_{e,t}^{\mathrm{ambhum}}$ is added to avoid perfectly repeated
humidity cycles,
\begin{equation}
\varepsilon_{e,t}^{\mathrm{ambhum}}
\sim
\mathcal N\!\left(
0,
(\sigma^{\mathrm{ambhum}})^2
\right),
\end{equation}
where
\[
\sigma^{\mathrm{ambhum}}
=
2\text{\%}.
\]

The ambient relative humidity is calculated as
\begin{equation}
H_{e,t}^{\mathrm{amb}}
=
\min\!\left(
95,
\max\!\left(
30,
H_{e,\mathrm{base}}^{\mathrm{amb}}
+
10\sin\!\left(
\frac{2\pi d_t}{365}
\right)
+
5\sin\!\left(
\frac{2\pi(h_t+6)}{24}
\right)
+
\varepsilon_{e,t}^{\mathrm{ambhum}}
\right)
\right).
\end{equation}

The yearly variation can change humidity by up to 10\% around the reference value, while the daily variation can
change it by up to 5\%. The final value is limited to
\([30\%,95\%]\) to avoid extreme generated humidity values.

\subsection{Door Use and Door Seal}
\label{app:door-use}

Door use refers to the opening and closing of the refrigerator door.
Door openings affect the internal temperature and energy consumption
\cite{Qiao2025,BelmanFlores2024}, while their frequency also varies
during the day \cite{Kapici2022}. Therefore, the simulator generates
door openings randomly using a probability that depends on the time of
day and a duration that can vary between openings.

In addition to normal door use, the simulator accounts for door-seal
degradation. The door seal limits air exchange between the refrigerated
compartment and the ambient air when the door is closed, whereas a
degraded seal increases heat and moisture exchange. The simulator
approximates this effect through the seal-efficiency factor
$\eta_{e,t}^{\mathrm{seal}}$ and short seal-related openings.

\paragraph{Door state variables:}
Four components of $Z_{e,t}$ describe the door. The binary state
$s_{e,t}^{\mathrm{door}}\in\{0,1\}$ indicates whether the door is
closed or open. The value $L_{e,t}^{\mathrm{door}}$ gives the number
of minutes since the current opening started,
$q_{e,t}^{\mathrm{door}}$ stores its closing time, and
$C_{e,t}^{\mathrm{door}}$ counts the openings since the beginning of
the simulation.
\[
s_{e,0}^{\mathrm{door}}=0,
\qquad
L_{e,0}^{\mathrm{door}}=0,
\qquad
q_{e,0}^{\mathrm{door}}=-1,
\qquad
C_{e,0}^{\mathrm{door}}=0.
\]
The value $q_{e,0}^{\mathrm{door}}=-1$ means that no closing time has
been scheduled yet.

\paragraph{Probability of a normal opening:}
When the door is closed, the simulator first determines the
probability of a normal user opening. Since door-opening frequency
varies throughout the day \cite{Kapici2022}, $\theta_e$ contains five
time-dependent probabilities, namely $p_e^{\mathrm{night}}$,
$p_e^{\mathrm{morning}}$, $p_e^{\mathrm{noon}}$,
$p_e^{\mathrm{evening}}$, and $p_e^{\mathrm{other}}$. The probability
used at time $t$ depends on the current hour $h_t$ and is given by
\begin{equation}
p_{e,t}^{\mathrm{door}} =
\begin{cases}
p_e^{\mathrm{night}}, & 0 \le h_t < 6,\\
p_e^{\mathrm{morning}}, & 6 \le h_t < 9,\\
p_e^{\mathrm{noon}}, & 11 \le h_t < 14,\\
p_e^{\mathrm{evening}}, & 17 \le h_t < 20,\\
p_e^{\mathrm{other}}, & \text{otherwise}.
\end{cases}
\end{equation}

With this probability set, the simulator decides whether an
opening actually starts. It generates the binary variable
$\zeta_{e,t}^{\mathrm{door}}\in\{0,1\}$ according to
\begin{equation}
\zeta_{e,t}^{\mathrm{door}}
\sim
\operatorname{Bernoulli}
\left(
p_{e,t}^{\mathrm{door}}
\right).
\end{equation}
An opening is generated when
$\zeta_{e,t}^{\mathrm{door}}=1$.

\paragraph{Duration of an opening:}
When a normal opening is generated, its duration is obtained from
\begin{equation}
\xi_{e,t}^{\mathrm{door}}
\sim
\mathcal N
\left(
\mu^{\mathrm{door}},
(\sigma^{\mathrm{door}})^2
\right),
\qquad
\mu^{\mathrm{door}}=0.5,
\qquad
\sigma^{\mathrm{door}}=0.8.
\end{equation}

The resulting duration is
\begin{equation}
D_{e,t}^{\mathrm{door}}
=
\begin{cases}
\displaystyle
\min\!\left(
10,\,
\max\!\left(
1,\,
\left\lfloor e^{\xi_{e,t}^{\mathrm{door}}}\right\rfloor
\right)
\right),
& \zeta_{e,t}^{\mathrm{door}}=1,
\\[2mm]
0,
& \zeta_{e,t}^{\mathrm{door}}=0.
\end{cases}
\end{equation}

\paragraph{Door-seal efficiency:}
The door-seal efficiency is given by,
\begin{equation}
\eta_{e,t}^{\mathrm{seal}}
=
\begin{cases}
\displaystyle
\max\!\left(
0.70,\,
1-0.30P_{e,j,t}
\right),
&
t_{e,j}^{\mathrm{deg}}
\le t <
t_{e,j}^{\mathrm{fail}},
\quad
\text{for a \texttt{DOOR\_SEAL} degradation},
\\[2mm]
1,
& \text{otherwise}.
\end{cases}
\end{equation}
The degradation progress \(P_{e,j,t}\) $\in [0,1]$ is defined in
Eq.~\eqref{eq:degradation-progress-appendix}. The minimum value of
0.70 keeps the seal partially effective while allowing its condition
to progressively deteriorate before failure.

\paragraph{Seal-triggered openings:}
When the door is closed and the seal efficiency falls below 0.9, the
simulator can generate an additional opening associated with seal
degradation,
\begin{equation}
\zeta_{e,t}^{\mathrm{seal}}
=
\begin{cases}
\displaystyle
\operatorname{Bernoulli}
\left(
(1-\eta_{e,t}^{\mathrm{seal}})10^{-3}
\right),
&
s_{e,t}^{\mathrm{door}}=0
\text{ and }
\eta_{e,t}^{\mathrm{seal}}<0.9,
\\[2mm]
0,
& \text{otherwise}.
\end{cases}
\end{equation}
The threshold of 0.9 activates this effect only after the seal has
degraded, while the factor $10^{-3}$ keeps these additional openings
rare. Their probability increases as the seal efficiency decreases.

The duration actually used by the simulator,
$D_{e,t}^{\mathrm{used}}$, is
\begin{equation}
D_{e,t}^{\mathrm{used}}
=
\begin{cases}
\displaystyle
\max\!\left(
1,\,
\left\lfloor
0.3D_{e,t}^{\mathrm{door}}
\right\rfloor
\right),
&
\zeta_{e,t}^{\mathrm{seal}}=1,
\\[2mm]
D_{e,t}^{\mathrm{door}},
&
\zeta_{e,t}^{\mathrm{seal}}=0.
\end{cases}
\end{equation}
The factor 0.3 keeps seal-related openings shorter than normal user
openings, with a minimum duration of one minute.

\paragraph{Applying the opening:}
When the used opening duration $D_{e,t}^{\mathrm{used}}>0$, the door
opens and the door state becomes
$s_{e,t}^{\mathrm{door}}=1$. The closing time
$q_{e,t}^{\mathrm{door}}$ is then
\begin{equation}
q_{e,t}^{\mathrm{door}}
=
t+D_{e,t}^{\mathrm{used}}.
\end{equation}

The opening counter $C_{e,t}^{\mathrm{door}}$ increases by 1, while
the door-open duration $L_{e,t}^{\mathrm{door}}$ counts the number of
minutes since the opening started. When
$t=q_{e,t}^{\mathrm{door}}$, the door closes, the door state returns
to 0, and the door-open duration is reset,
\begin{equation}
s_{e,t}^{\mathrm{door}}=0,
\qquad
L_{e,t}^{\mathrm{door}}=0.
\end{equation}

\subsection{Defrost State and Heater Efficiency}
\label{app:defrost-state}

Frost can accumulate on the evaporator during refrigerator operation.
Many frost-free refrigerators use repeated defrost cycles to heat the
evaporator and remove this frost \cite{Braun2025,Nidamanuri2026}.
The simulator follows this principle by scheduling repeated defrost
cycles. It also allows the defrost-heater efficiency to decrease during
a \texttt{DEFROST\_FAIL} degradation.

\paragraph{Planned defrost schedule:}
Before the simulation begins, the simulator creates the planned
defrost times for refrigerator $e$. Four components of $\theta_e$
define this schedule. The first reference minute is
$m_{e,\mathrm{first}}^{\mathrm{def}}$, the interval between
consecutive reference starts is $\Delta_e^{\mathrm{def}}$, the
duration of a cycle is $L_e^{\mathrm{def}}$, and the maximum shift
from a reference start is $J_e^{\mathrm{def}}$. All four quantities
are expressed in minutes.

For each defrost cycle $k$, the simulator generates a random shift
$U_{e,k}^{\mathrm{def}}$,
\begin{equation}
U_{e,k}^{\mathrm{def}}
\sim
\mathcal U_{\mathrm{disc}}
\left(
-J_e^{\mathrm{def}},
J_e^{\mathrm{def}}
\right).
\label{eq:defrost-start-shift}
\end{equation}
A negative shift starts the cycle earlier than its reference time,
while a positive shift starts it later. The planned start time
$\bar t_{e,k}^{\mathrm{def}}$ is
\begin{equation}
\bar t_{e,k}^{\mathrm{def}}
=
1
+
m_{e,\mathrm{first}}^{\mathrm{def}}
+
(k-1)\Delta_e^{\mathrm{def}}
+
U_{e,k}^{\mathrm{def}}.
\label{eq:defrost-planned-start}
\end{equation}

Planned cycles are generated while their reference starts remain
within the generated time series. Their start times are stored in
$\mathcal D_e$,
\begin{equation}
\mathcal D_e
=
\left\{
\bar t_{e,k}^{\mathrm{def}}
\;\middle|\;
m_{e,\mathrm{first}}^{\mathrm{def}}
+
(k-1)\Delta_e^{\mathrm{def}}
<
T_e
\right\}.
\label{eq:defrost-schedule}
\end{equation}

\paragraph{Defrost state during simulation:}
Two components of $Z_{e,t}$ describe the current defrost cycle. The
defrost state $s_{e,t}^{\mathrm{def}}\in\{0,1\}$ indicates whether
defrost is inactive or active, while the ending time
$q_{e,t}^{\mathrm{def}}$ stores when the active cycle must end. Their
initial values are
\begin{equation}
s_{e,0}^{\mathrm{def}}=0,
\qquad
q_{e,0}^{\mathrm{def}}=-1.
\end{equation}
The initial value $q_{e,0}^{\mathrm{def}}=-1$ indicates that no ending
time has yet been scheduled.

When the defrost state is inactive, a cycle starts when the current
time matches a planned start time,
\begin{equation}
t\in\mathcal D_e.
\label{eq:defrost-start-condition}
\end{equation}
The defrost state then becomes active and the ending time is set from
the configured cycle duration,
\begin{equation}
s_{e,t}^{\mathrm{def}}=1,
\qquad
q_{e,t}^{\mathrm{def}}
=
t+L_e^{\mathrm{def}}.
\label{eq:defrost-state-start}
\end{equation}
When the current time reaches the ending time
$q_{e,t}^{\mathrm{def}}$, the defrost state returns to 0.

\paragraph{Heater efficiency:}
The defrost heater heats the evaporator to remove accumulated frost.
Its efficiency is described by the heater-efficiency factor
$\eta_{e,t}^{\mathrm{def}}\in[0.30,1]$, where 1 corresponds to nominal
heater operation. Its value is
\begin{equation}
\eta_{e,t}^{\mathrm{def}}
=
\begin{cases}
\displaystyle
\max\!\left(
0.30,\,
1-0.70P_{e,j,t}
\right),
&
t_{e,j}^{\mathrm{deg}}
\leq t <
t_{e,j}^{\mathrm{fail}},
\quad
\text{for a \texttt{DEFROST\_FAIL} degradation},
\\[2mm]
1,
& \text{otherwise}.
\end{cases}
\label{eq:defrost-heater-efficiency}
\end{equation}
The degradation progress $P_{e,j,t}$ is defined in
Eq.~\eqref{eq:degradation-progress-appendix}. The lower bound of 0.30
keeps the heater partially effective while allowing a substantial
decrease before failure. The heater-efficiency factor is later used
to calculate defrost-heater current and frost removal.

\subsection{Thermostat and Compressor State}
\label{app:thermostat-compressor}

The thermostat controls when the compressor starts and stops according
to the temperatures inside the refrigerator. In a single-speed
refrigerator, the compressor is either active or inactive. It starts
when a compartment becomes too warm and stops when the compartment has
cooled sufficiently \cite{Bao2025,Caglayan2022}. The simulator follows
this principle using a lower and an upper temperature threshold for
each available compartment.

\paragraph{Thermostat thresholds:}
The thermostat thresholds are components of $\theta_e$. For the fridge,
they are $\tau_{e,\mathrm{fr}}^{\mathrm{low}}$ and
$\tau_{e,\mathrm{fr}}^{\mathrm{high}}$, and for the freezer they are
$\tau_{e,\mathrm{fz}}^{\mathrm{low}}$ and
$\tau_{e,\mathrm{fz}}^{\mathrm{high}}$. All thresholds are expressed
in $^{\circ}\mathrm{C}$. The upper threshold starts the compressor when
a compartment becomes too warm, while the lower threshold stops it
after sufficient cooling. Only compartments with configured thresholds
are included in the decision.

\paragraph{Compressor state:}
The compressor state
$s_{e,t}^{\mathrm{comp}}\in\{0,1\}$ is a component of $Z_{e,t}$,
where 0 indicates that the compressor is inactive and 1 that it is
active.

\paragraph{Ambient adjustment of the fridge threshold:}
Ambient temperature affects refrigerator cooling demand
\cite{ChenLi2022}. The simulator therefore adjusts the upper fridge
threshold according to the ambient temperature
$T_{e,t}^{\mathrm{amb}}$ (see \ref{app:ambient-conditions} in the Appendix), while the lower threshold
remains unchanged,
\begin{equation}
\widetilde{\tau}_{e,\mathrm{fr},t}^{\mathrm{high}}
=
\begin{cases}
\tau_{e,\mathrm{fr}}^{\mathrm{high}}-0.5,
& T_{e,t}^{\mathrm{amb}}>25,\\
\tau_{e,\mathrm{fr}}^{\mathrm{high}}+0.5,
& T_{e,t}^{\mathrm{amb}}<18,\\
\tau_{e,\mathrm{fr}}^{\mathrm{high}},
& \text{otherwise}.
\end{cases}
\label{eq:fridge-adjusted-threshold}
\end{equation}
This adjustment makes the compressor start at a lower fridge
temperature under warmer ambient conditions and at a higher
temperature under cooler conditions.

\paragraph{Ambient adjustment of the freezer threshold:}
The freezer upper threshold is also adjusted according to ambient
temperature,
\begin{equation}
\widetilde{\tau}_{e,\mathrm{fz},t}^{\mathrm{high}}
=
\begin{cases}
\tau_{e,\mathrm{fz}}^{\mathrm{high}}+1,
& T_{e,t}^{\mathrm{amb}}>25,\\
\tau_{e,\mathrm{fz}}^{\mathrm{high}}-1,
& T_{e,t}^{\mathrm{amb}}<18,\\
\tau_{e,\mathrm{fz}}^{\mathrm{high}},
& \text{otherwise}.
\end{cases}
\label{eq:freezer-adjusted-threshold}
\end{equation}
A larger adjustment of $1\,^{\circ}\mathrm{C}$ is used for the freezer
to distinguish its response from that of the fridge compartment.

\paragraph{Compressor switching rule:}
The defrost state $s_{e,t}^{\mathrm{def}}$ (see \ref{app:defrost-state} in the Appendix) is checked first because the
compressor remains inactive during defrost
\cite{ChenLi2022}. Outside defrost, the
compartment temperatures are compared with their thermostat thresholds
before the thermal update of the current minute. The compressor state
is
\begin{equation}
s_{e,t}^{\mathrm{comp}}
=
\begin{cases}
0,
& t=0,\\
0,
& s_{e,t}^{\mathrm{def}}=1,\\
1,
& \text{at least one available compartment is above its adjusted
upper threshold},\\
0,
& \text{all available compartments are below their lower thresholds},\\
s_{e,t-1}^{\mathrm{comp}},
& \text{otherwise}.
\end{cases}
\label{eq:compressor-hysteresis}
\end{equation}
When temperatures remain between the lower and upper thresholds, the
compressor keeps its previous state, preventing frequent switching
caused by small temperature changes.

\paragraph{Compressor duty cycle:}
The compressor duty cycle $DC_{e,t}$ is the percentage of a recent
period during which the compressor was active \cite{ChenLi2022}. The
history length $H_e^{\mathrm{hist}}$, in minutes, is a component of
$\theta_e$. Compressor states before the first observation are taken
as zero,
\begin{equation}
s_{e,t'}^{\mathrm{comp}}=0,
\qquad
t'\leq0.
\end{equation}
The duty cycle is then
\begin{equation}
DC_{e,t}
=
\frac{100}{H_e^{\mathrm{hist}}}
\sum_{\kappa=0}^{H_e^{\mathrm{hist}}-1}
s_{e,t-\kappa}^{\mathrm{comp}}.
\label{eq:compressor-duty-cycle}
\end{equation}
The sum counts the active compressor minutes over the most recent
$H_e^{\mathrm{hist}}$ minutes, and the factor 100 expresses this
fraction as a percentage.

\subsection{Thermal Behaviour}
\label{app:thermal-behaviour}

At each instant $t$, the simulator updates four temperatures, which
are components of $Z_{e,t}$: the fridge temperature
$T_{e,t}^{\mathrm{fr}}$, the freezer temperature
$T_{e,t}^{\mathrm{fz}}$, the evaporator temperature
$T_{e,t}^{\mathrm{ev}}$, and the condenser temperature
$T_{e,t}^{\mathrm{cd}}$. All four are expressed in
$^{\circ}\mathrm{C}$. Their initial values are specified in
$\theta_e$, which also contains the fridge and freezer target
temperatures $T_e^{\mathrm{fr,target}}$ and
$T_e^{\mathrm{fz,target}}$ when the corresponding compartments are
present.

These temperatures change with compressor operation, airflow, door
openings, and defrost. Simplified refrigerator models use these
relationships to calculate temperature changes \cite{Yoo2024}. Door
openings can increase compartment temperatures
\cite{BelmanFlores2024}, while defrost changes evaporator and
compartment temperatures \cite{Nidamanuri2025,Braun2025}.

The degradation progress $P_{e,j,t}$ used below is defined in
Eq.~\eqref{eq:degradation-progress-appendix}.

\paragraph{Compressor efficiency:}
The compressor-efficiency factor
$\eta_{e,t}^{\mathrm{comp}}$ is a component of $Z_{e,t}$. A value of
1 corresponds to nominal compressor operation, while lower values
reduce the cooling effect. Refrigerant loss can reduce evaporator
cooling capacity and change compartment temperatures
\cite{ZhangLeakage2024}. The compressor efficiency is
\begin{equation}
\eta_{e,t}^{\mathrm{comp}}
=
\begin{cases}
1,
& t=0,
\\[2mm]
\displaystyle
\max\!\left(
0.5,\,
1-0.5P_{e,j,t}^{1.5}
\right),
& \text{during a \texttt{COMP\_DEGRAD} degradation},
\\[2mm]
\displaystyle
\max\!\left(
0.85,\,
1-0.15P_{e,j,t}
\right),
& \text{during a \texttt{COMP\_WEAR} degradation},
\\[2mm]
\eta_{e,t-1}^{\mathrm{comp}},
& \text{otherwise}.
\end{cases}
\label{eq:compressor-efficiency-degradation}
\end{equation}
The lower bound of 0.5 under \texttt{COMP\_DEGRAD} produces a clear
loss of cooling performance before failure. The smaller decrease to
0.85 under \texttt{COMP\_WEAR} distinguishes mechanical wear from the
stronger cooling loss produced by \texttt{COMP\_DEGRAD}.

\paragraph{Compressor mechanical wear:}
The mechanical-wear factor $w_{e,t}^{\mathrm{mec}}\in[0,1]$ is a
component of $Z_{e,t}$, where 0 corresponds to no wear and larger
values indicate increasing wear. Its value is
\begin{equation}
w_{e,t}^{\mathrm{mec}}
=
\begin{cases}
w_{e,0}^{\mathrm{mec}},
& t=0,
\\[2mm]
\displaystyle
\min\!\left(
w_{e,0}^{\mathrm{mec}}+P_{e,j,t},
1
\right),
& \text{during a \texttt{COMP\_WEAR} degradation},
\\[2mm]
w_{e,t-1}^{\mathrm{mec}},
& \text{otherwise}.
\end{cases}
\label{eq:mechanical-wear-degradation}
\end{equation}

\paragraph{Condenser fouling:}
Condenser fouling can affect refrigerator temperature and energy
performance \cite{PardoCely2023}. The thermal-resistance factor
$R_{e,t}^{\mathrm{th}}$ is a component of $Z_{e,t}$, where larger
values indicate greater difficulty rejecting heat. Its initial value
$R_{e,0}^{\mathrm{th}}$ is specified in $\theta_e$ and is equal to 1
when no initial increase in thermal resistance is configured. Its
value is
\begin{equation}
R_{e,t}^{\mathrm{th}}
=
\begin{cases}
R_{e,0}^{\mathrm{th}},
& t=0,
\\[2mm]
R_{e,0}^{\mathrm{th}}
+
0.6P_{e,j,t}^{0.8},
& \text{during a \texttt{COND\_DIRT} degradation},
\\[2mm]
R_{e,t-1}^{\mathrm{th}},
& \text{otherwise}.
\end{cases}
\label{eq:condenser-resistance-degradation}
\end{equation}
The coefficient 0.6 produces a progressive increase in thermal
resistance as condenser degradation develops.

\paragraph{Fan efficiency:}
The fan-efficiency factor $\eta_{e,t}^{\mathrm{fan}}$ describes the
effect of fan airflow on fridge cooling \cite{Yoo2024} and is a
component of $Z_{e,t}$. Its value is
\begin{equation}
\eta_{e,t}^{\mathrm{fan}}
=
\begin{cases}
1,
& t=0,
\\[2mm]
\displaystyle
\max\!\left(
0.60,\,
1-0.40P_{e,j,t}
\right),
& \text{during a \texttt{FAN\_DEGRAD} degradation},
\\[2mm]
\eta_{e,t-1}^{\mathrm{fan}},
& \text{otherwise}.
\end{cases}
\label{eq:fan-efficiency-degradation}
\end{equation}
The lower bound of 0.60 produces a progressive loss of fan performance
while keeping the fan partially operational before failure.

Each temperature update includes a small random variation, expressed in $^{\circ}\mathrm{C}$, to avoid
perfectly smooth temperature curves. The compartment temperatures use
smaller variations, while larger short-term variations are assigned for
the evaporator and condenser temperatures. The variations are
\begin{align}
\nu_{e,t}^{\mathrm{fr}}
&\sim
\mathcal N\!\left(0,(\sigma^{\mathrm{fr}})^2\right),
&
\nu_{e,t}^{\mathrm{fz}}
&\sim
\mathcal N\!\left(0,(\sigma^{\mathrm{fz}})^2\right),
\\
\nu_{e,t}^{\mathrm{ev}}
&\sim
\mathcal N\!\left(0,(\sigma^{\mathrm{ev}})^2\right),
&
\nu_{e,t}^{\mathrm{cd}}
&\sim
\mathcal N\!\left(0,(\sigma^{\mathrm{cd}})^2\right),
\\
\nu_{e,t}^{\mathrm{cd,on}}
&\sim
\mathcal N\!\left(0,(\sigma^{\mathrm{cd,on}})^2\right),
\end{align}
with
\[
\sigma^{\mathrm{fr}}
=
\sigma^{\mathrm{fz}}
=
0.1,
\qquad
\sigma^{\mathrm{ev}}
=
0.2,
\qquad
\sigma^{\mathrm{cd}}
=
0.5,
\qquad
\sigma^{\mathrm{cd,on}}
=
2,
\]

\paragraph{Evaporator temperature:}
The evaporator temperature is updated first because its cooling and
warming behaviour affects the other temperature updates. When the
compressor is active, the evaporator moves toward a low target
temperature. This target is set $5\,^{\circ}\mathrm{C}$ below the
freezer target when a freezer is present, keeping the evaporator colder
than the compartment it cools. When no freezer is present, a reference
target of $-25\,^{\circ}\mathrm{C}$ is used,
\begin{equation}
T_e^{\mathrm{ev,target}}
=
\begin{cases}
T_e^{\mathrm{fz,target}}-5,
& \text{if a freezer compartment is present},\\
-25,
& \text{otherwise}.
\end{cases}
\label{eq:evaporator-target}
\end{equation}

When the compressor is inactive, the evaporator moves toward the
temperature of the refrigerated compartment. The freezer temperature
is used when a freezer is present, and the fridge temperature
otherwise,
\begin{equation}
T_{e,t-1}^{\mathrm{ev,ref}}
=
\begin{cases}
T_{e,t-1}^{\mathrm{fz}},
& \text{if a freezer compartment is present},\\
T_{e,t-1}^{\mathrm{fr}},
& \text{otherwise}.
\end{cases}
\label{eq:evaporator-reference-temperature}
\end{equation}

Using the defrost state $s_{e,t}^{\mathrm{def}}$
(see \ref{app:defrost-state} in the Appendix) and the compressor state
$s_{e,t}^{\mathrm{comp}}$
(see \ref{app:thermostat-compressor} in the Appendix), the evaporator
temperature is updated as
\begin{equation}
T_{e,t}^{\mathrm{ev}}
=
\begin{cases}
\min\!\left(
10,\,
T_{e,t-1}^{\mathrm{ev}}+2
\right)
+
\nu_{e,t}^{\mathrm{ev}},
&
s_{e,t}^{\mathrm{def}}=1,
\\[2mm]
T_{e,t-1}^{\mathrm{ev}}
+
0.1\eta_{e,t}^{\mathrm{comp}}
\left(
T_e^{\mathrm{ev,target}}
-
T_{e,t-1}^{\mathrm{ev}}
\right)
+
\nu_{e,t}^{\mathrm{ev}},
&
s_{e,t}^{\mathrm{def}}=0,\;
s_{e,t}^{\mathrm{comp}}=1,
\\[2mm]
T_{e,t-1}^{\mathrm{ev}}
+
0.05
\left(
T_{e,t-1}^{\mathrm{ev,ref}}
-
T_{e,t-1}^{\mathrm{ev}}
\right)
+
\nu_{e,t}^{\mathrm{ev}},
&
s_{e,t}^{\mathrm{def}}=0,\;
s_{e,t}^{\mathrm{comp}}=0.
\end{cases}
\label{eq:evaporator-temperature-update}
\end{equation}

During defrost, a change of $2\,^{\circ}\mathrm{C}$ per minute allows
the evaporator to warm rapidly, while the $10\,^{\circ}\mathrm{C}$
limit prevents an excessive temperature increase. When the compressor
is active, the coefficient 0.1 produces a faster movement toward the
cooling target, with this effect reduced by compressor degradation.
When the compressor is inactive, the smaller coefficient 0.05 gives a
slower return toward the compartment temperature.

\paragraph{Condenser temperature:}
The condenser rejects heat to the ambient air, so its temperature
depends on the ambient temperature $T_{e,t}^{\mathrm{amb}}$, the
compressor state $s_{e,t}^{\mathrm{comp}}$, the thermal-resistance
factor $R_{e,t}^{\mathrm{th}}$, and the compressor efficiency
$\eta_{e,t}^{\mathrm{comp}}$. The condenser temperature is updated as
\begin{equation}
T_{e,t}^{\mathrm{cd}}
=
\begin{cases}
T_{e,t}^{\mathrm{amb}}
+
15R_{e,t}^{\mathrm{th}}
+
8
\left(
1-\eta_{e,t}^{\mathrm{comp}}
\right)
+
\nu_{e,t}^{\mathrm{cd,on}}
+
\nu_{e,t}^{\mathrm{cd}},
&
s_{e,t}^{\mathrm{comp}}=1,
\\[2mm]
T_{e,t-1}^{\mathrm{cd}}
+
0.1
\left(
T_{e,t}^{\mathrm{amb}}
-
T_{e,t-1}^{\mathrm{cd}}
\right)
+
\nu_{e,t}^{\mathrm{cd}},
&
s_{e,t}^{\mathrm{comp}}=0.
\end{cases}
\label{eq:condenser-temperature-update}
\end{equation}

When the compressor is active, the factor 15 keeps the condenser
temperature clearly above ambient under normal conditions, while the
factor 8 adds a moderate temperature increase as compressor efficiency
decreases. When the compressor is inactive, the coefficient 0.1 gives
a gradual return toward ambient temperature rather than an immediate
change.

\paragraph{Freezer temperature:}
The freezer temperature depends on the defrost state
$s_{e,t}^{\mathrm{def}}$, the compressor state
$s_{e,t}^{\mathrm{comp}}$, the door state
$s_{e,t}^{\mathrm{door}}$, and the door-seal efficiency
$\eta_{e,t}^{\mathrm{seal}}$. The defrost and compressor states are
defined in Sections~\ref{app:defrost-state} and
\ref{app:thermostat-compressor} of the Appendix, while the door quantities are defined
in Section~\ref{app:door-use} of the Appendix. The freezer temperature is updated as
\begin{equation}
T_{e,t}^{\mathrm{fz}}
=
\begin{cases}
\min\!\left(
-4,\,
T_{e,t-1}^{\mathrm{fz}}+0.8
\right)
+
\nu_{e,t}^{\mathrm{fz}},
&
s_{e,t}^{\mathrm{def}}=1,
\\[2mm]
\max\!\left(
-35,\,
T_{e,t-1}^{\mathrm{fz}}
-
0.4\eta_{e,t}^{\mathrm{comp}}
\right)
+
\nu_{e,t}^{\mathrm{fz}},
&
s_{e,t}^{\mathrm{def}}=0,\;
s_{e,t}^{\mathrm{comp}}=1,
\\[2mm]
T_{e,t-1}^{\mathrm{fz}}
+
0.16
\left[
1+
1.5
\left(
1-\eta_{e,t}^{\mathrm{seal}}
\right)
\right]
+
\nu_{e,t}^{\mathrm{fz}},
&
s_{e,t}^{\mathrm{def}}=0,\;
s_{e,t}^{\mathrm{comp}}=0,\;
s_{e,t}^{\mathrm{door}}=1,
\\[2mm]
T_{e,t-1}^{\mathrm{fz}}
+
0.08
\left[
1+
1.5
\left(
1-\eta_{e,t}^{\mathrm{seal}}
\right)
\right]
+
\nu_{e,t}^{\mathrm{fz}},
&
s_{e,t}^{\mathrm{def}}=0,\;
s_{e,t}^{\mathrm{comp}}=0,\;
s_{e,t}^{\mathrm{door}}=0.
\end{cases}
\label{eq:freezer-temperature-update}
\end{equation}

During defrost, the $0.8\,^{\circ}\mathrm{C}$ increase allows the
freezer to warm gradually, while the $-4\,^{\circ}\mathrm{C}$ limit
prevents excessive warming. During cooling, the
$0.4\,^{\circ}\mathrm{C}$ decrease provides progressive cooling, and
the $-35\,^{\circ}\mathrm{C}$ limit prevents unrealistically low
temperatures. When the compressor is inactive, the freezer warms twice
as fast with the door open as with the door closed. The factor 1.5
increases this warming as the door seal degrades.

\paragraph{Fridge temperature:}
The fridge temperature follows the same general logic as the freezer,
using the defrost state $s_{e,t}^{\mathrm{def}}$, the compressor state
$s_{e,t}^{\mathrm{comp}}$, the door state
$s_{e,t}^{\mathrm{door}}$, and the door-seal efficiency
$\eta_{e,t}^{\mathrm{seal}}$. The fan efficiency
$\eta_{e,t}^{\mathrm{fan}}$ also affects cooling when the compressor
is active. The fridge temperature is updated as
\begin{equation}
T_{e,t}^{\mathrm{fr}}
=
\begin{cases}
\min\!\left(
8,\,
T_{e,t-1}^{\mathrm{fr}}+0.5
\right)
+
\nu_{e,t}^{\mathrm{fr}},
&
s_{e,t}^{\mathrm{def}}=1,
\\[2mm]
\max\!\left(
-2,\,
T_{e,t-1}^{\mathrm{fr}}
-
0.25
\eta_{e,t}^{\mathrm{comp}}
\eta_{e,t}^{\mathrm{fan}}
\right)
+
\nu_{e,t}^{\mathrm{fr}},
&
s_{e,t}^{\mathrm{def}}=0,\;
s_{e,t}^{\mathrm{comp}}=1,
\\[2mm]
T_{e,t-1}^{\mathrm{fr}}
+
0.30
\left[
1+
2
\left(
1-\eta_{e,t}^{\mathrm{seal}}
\right)
\right]
+
\nu_{e,t}^{\mathrm{fr}},
&
s_{e,t}^{\mathrm{def}}=0,\;
s_{e,t}^{\mathrm{comp}}=0,\;
s_{e,t}^{\mathrm{door}}=1,
\\[2mm]
T_{e,t-1}^{\mathrm{fr}}
+
0.06
\left[
1+
2
\left(
1-\eta_{e,t}^{\mathrm{seal}}
\right)
\right]
+
\nu_{e,t}^{\mathrm{fr}},
&
s_{e,t}^{\mathrm{def}}=0,\;
s_{e,t}^{\mathrm{comp}}=0,\;
s_{e,t}^{\mathrm{door}}=0.
\end{cases}
\label{eq:fridge-temperature-update}
\end{equation}

During defrost, the $0.5\,^{\circ}\mathrm{C}$ increase gives gradual
warming, while the $8\,^{\circ}\mathrm{C}$ limit prevents excessive
temperature rise. During cooling, the $0.25\,^{\circ}\mathrm{C}$ rate
produces a progressive decrease, and the $-2\,^{\circ}\mathrm{C}$
limit prevents unrealistically low fridge temperatures. When the
compressor is inactive, the $0.30\,^{\circ}\mathrm{C}$ rate makes the
fridge warm five times faster with the door open than with the door
closed. The factor 2 increases this warming as the door seal degrades,
making its effect more visible before failure.

\subsection{Electrical Quantities and Fan Speed}
\label{app:electrical-quantities}

Electrical quantities describe the electricity used by the
refrigerator at each instant $t$. The compressor is a major electrical
load, while the fan and defrost heater also contribute to refrigerator
energy consumption \cite{BelmanFlores2024,Nidamanuri2026}. The light
adds another electrical load when the door is open. For a fixed supply
voltage, higher current produces higher electrical power.

The simulator also calculates fan speed because fan operation affects
cooling performance \cite{PardoCely2023}. In addition, the fan is a
source of refrigerator vibration and acoustic noise
\cite{Zarate2021}. The fan speed is therefore also used in the
vibration and acoustic calculations described in
Section~\ref{app:vibration-noise} of the Appendix.

\paragraph{Compressor current:}
The compressor current $I_{e,t}^{\mathrm{comp}}$ is based on the
reference current $I_e^{\mathrm{base}}$, in amperes, taken from
manufacturer data \cite{WhirlpoolTechW11529760,SecopSC12G}. The
reference current is first adjusted according to cooling demand and
then according to compressor and condenser conditions.

Compressor electrical demand can vary with refrigerator cooling
conditions \cite{BelmanFlores2024}. In the simulator, cooling demand
is estimated from the difference between the fridge temperature
$T_{e,t}^{\mathrm{fr}}$ and its target
$T_e^{\mathrm{fr,target}}$. The load factor is
\begin{equation}
\lambda_{e,t}
=
\min\!\left(
1,\,
\max\!\left(
0.3,\,
\frac{
T_{e,t}^{\mathrm{fr}}
-
T_e^{\mathrm{fr,target}}
}{
\Delta T^{\mathrm{mod}}
}
\right)
\right).
\label{eq:compressor-load-factor}
\end{equation}
where
\[
\Delta T^{\mathrm{mod}}=4\,^{\circ}\mathrm{C}.
\]
This value makes the load reach its maximum over a moderate temperature
difference. The lower bound of 0.3 keeps a minimum load when the
compressor is active, while the upper bound of 1 limits the load to
its maximum value.

The demand-adjusted current is
\begin{equation}
I_{e,t}^{\mathrm{base},*}
=
I_e^{\mathrm{base}}\lambda_{e,t}.
\label{eq:demand-adjusted-current}
\end{equation}

Compressor and condenser conditions can also affect compressor current
\cite{Silveira2023,KoJeong2024,PardoCely2023}. The simulator accounts
for the compressor efficiency $\eta_{e,t}^{\mathrm{comp}}$,
mechanical wear $w_{e,t}^{\mathrm{mec}}$, condenser thermal resistance
$R_{e,t}^{\mathrm{th}}$, and condenser temperature
$T_{e,t}^{\mathrm{cd}}$ through four multiplying factors,
\begin{align}
M_{e,t}^{\mathrm{eff}}
&=
1
+
0.8
\left(
\frac{1}{\eta_{e,t}^{\mathrm{comp}}}
-
1
\right),
\label{eq:current-factor-eff}
\\
M_{e,t}^{\mathrm{mec}}
&=
1
+
0.6w_{e,t}^{\mathrm{mec}},
\label{eq:current-factor-mec}
\\
M_{e,t}^{\mathrm{th}}
&=
1
+
0.4
\left(
R_{e,t}^{\mathrm{th}}-1
\right),
\label{eq:current-factor-th}
\\
M_{e,t}^{\mathrm{cd}}
&=
1
+
\alpha^{\mathrm{cd}}
\left(
T_{e,t}^{\mathrm{cd}}
-
T^{\mathrm{cd,ref}}
\right).
\label{eq:current-factor-cd}
\end{align}

The coefficients 0.8, 0.6, and 0.4 give compressor efficiency the
strongest effect on current, followed by mechanical wear and condenser
thermal resistance. For the condenser-temperature factor,
\[
T^{\mathrm{cd,ref}}
=
30\,^{\circ}\mathrm{C},
\qquad
\alpha^{\mathrm{cd}}
=
0.02\,^{\circ}\mathrm{C}^{-1}.
\]
The reference temperature of $30\,^{\circ}\mathrm{C}$ is consistent
with condensing conditions studied for household refrigerators
\cite{KimKim2014}. The coefficient
$0.02\,^{\circ}\mathrm{C}^{-1}$ produces a gradual current change as
the condenser temperature moves away from this reference.

A small random variation is added to avoid perfectly regular current
values,
\begin{equation}
\varepsilon_{e,t}^{I}
\sim
\mathcal N\!\left(
0,
(\sigma^{I})^2
\right),
\label{eq:compressor-current-noise}
\end{equation}
where
\[
\sigma^{I}=0.1\,\mathrm{A}.
\]
This value keeps the random variation small relative to normal
compressor current.

Using the compressor state $s_{e,t}^{\mathrm{comp}}$
(see \ref{app:thermostat-compressor} in the Appendix), the final compressor
current is
\begin{equation}
I_{e,t}^{\mathrm{comp}}
=
\begin{cases}
0,
&
s_{e,t}^{\mathrm{comp}}=0,
\\[2mm]
\max\!\left[
0,\,
I_{e,t}^{\mathrm{base},*}
M_{e,t}^{\mathrm{eff}}
M_{e,t}^{\mathrm{mec}}
M_{e,t}^{\mathrm{th}}
M_{e,t}^{\mathrm{cd}}
+
\varepsilon_{e,t}^{I}
\right],
&
s_{e,t}^{\mathrm{comp}}=1.
\end{cases}
\label{eq:compressor-current}
\end{equation}

\paragraph{Fan speed:}
Fan speed describes how fast the refrigerator fan rotates to circulate
air. It is based on the reference speed $RPM_e^{\mathrm{base}}$, taken
from manufacturer data \cite{WhirlpoolTechW11529760}, and the
fan-efficiency factor $\eta_{e,t}^{\mathrm{fan}}$ defined in
Section~\ref{app:thermal-behaviour} of the Appendix.

A small random variation is added to avoid perfectly regular fan
speeds,
\begin{equation}
\varepsilon_{e,t}^{\mathrm{rpm}}
\sim
\mathcal N\!\left(
0,
(\sigma^{\mathrm{rpm}})^2
\right),
\label{eq:fan-speed-noise}
\end{equation}
where
\[
\sigma^{\mathrm{rpm}}=50\,\mathrm{rpm}.
\]
This value keeps the minute-to-minute variation small relative to the
reference fan speed.

Using the compressor state $s_{e,t}^{\mathrm{comp}}$ and the defrost
state $s_{e,t}^{\mathrm{def}}$, the fan speed is
\begin{equation}
RPM_{e,t}
=
\begin{cases}
RPM_e^{\mathrm{base}}
\eta_{e,t}^{\mathrm{fan}}
+
\varepsilon_{e,t}^{\mathrm{rpm}},
&
s_{e,t}^{\mathrm{comp}}=1,\;
s_{e,t}^{\mathrm{def}}=0,\;
\eta_{e,t}^{\mathrm{fan}}>0.1,
\\[2mm]
0,
& \text{otherwise}.
\end{cases}
\label{eq:fan-speed}
\end{equation}

The threshold of 0.1 prevents fan operation if its efficiency becomes
extremely low. Since $\eta_{e,t}^{\mathrm{fan}}$ remains at or above
0.60 in the generated data, this threshold acts only as a safeguard.

\paragraph{Defrost-heater current:}
The defrost heater uses electrical current to heat the evaporator and
remove frost during a defrost cycle
\cite{Nidamanuri2025,ZhangDefrost2025}. This operating principle is
also described in manufacturer technical data
\cite{WhirlpoolTechW11529760}.

The simulator uses the defrost state $s_{e,t}^{\mathrm{def}}$ and the
heater-efficiency factor $\eta_{e,t}^{\mathrm{def}}$, both defined in
Section~\ref{app:defrost-state} of the Appendix. The reference heater current is
$I^{\mathrm{def,base}}=3\,\mathrm{A}$
\cite{WhirlpoolTechW11529760}.

A small random variation is added to the heater current,
\begin{equation}
\varepsilon_{e,t}^{\mathrm{def}}
\sim
\mathcal N\!\left(
0,
(\sigma^{\mathrm{def}})^2
\right),
\label{eq:defrost-current-noise}
\end{equation}
where
\[
\sigma^{\mathrm{def}}=0.2\,\mathrm{A}.
\]

The heater current is
\begin{equation}
I_{e,t}^{\mathrm{def}}
=
\begin{cases}
I^{\mathrm{def,base}}
\eta_{e,t}^{\mathrm{def}}
+
\varepsilon_{e,t}^{\mathrm{def}},
&
s_{e,t}^{\mathrm{def}}=1,\;
\eta_{e,t}^{\mathrm{def}}>0.3,
\\[2mm]
0,
& \text{otherwise}.
\end{cases}
\label{eq:defrost-heater-current}
\end{equation}

The current decreases with heater efficiency, making the degradation
visible in the electrical output. The threshold of 0.30 treats a
severely degraded heater as failed, while the random variation
introduces small current changes during operation.

\paragraph{Electrical power:}
The total electrical power combines the contributions of the
compressor, defrost heater, fan, and light. The light uses the fixed
power $P_e^{\mathrm{light}}$ when the door is open, while the fan uses
the fixed power $P_e^{\mathrm{fan}}$ when it is running.

Because only the fan running state is needed for this calculation, the
fan speed $RPM_{e,t}$ is converted into the binary state
\begin{equation}
\chi_{e,t}^{\mathrm{fan}}
=
\mathbb{1}
\left\{
RPM_{e,t}>0
\right\},
\label{eq:fan-power-activity}
\end{equation}
where $\mathbb{1}\{\cdot\}$ equals 1 when the condition is true and 0
otherwise.

The supply voltage is denoted by $U_e$, in volts. Using the compressor
current $I_{e,t}^{\mathrm{comp}}$, the defrost-heater current
$I_{e,t}^{\mathrm{def}}$, and the door state
$s_{e,t}^{\mathrm{door}}$, the electrical power
$P_{e,t}^{\mathrm{elec}}$, in watts, is
\begin{equation}
P_{e,t}^{\mathrm{elec}}
=
\max\!\left[
0,\,
\left(
I_{e,t}^{\mathrm{comp}}
+
I_{e,t}^{\mathrm{def}}
\right)
U_e
+
P_e^{\mathrm{fan}}
\chi_{e,t}^{\mathrm{fan}}
+
P_e^{\mathrm{light}}
s_{e,t}^{\mathrm{door}}
\right].
\label{eq:electrical-power}
\end{equation}

The compressor and defrost-heater contributions are calculated from
current and supply voltage, while the fan and light use their fixed
power ratings. The outer maximum prevents negative power values. This
simplified calculation does not include power factor or motor-drive
losses.

\paragraph{Cumulative electrical energy:}
The cumulative electrical energy $E_{e,t}$, in kilowatt-hours, is a
component of $Z_{e,t}$. Since the simulation uses a one-minute time
step,
\[
\Delta t_h=\frac{1}{60}
\]
is the corresponding duration in hours. The cumulative energy is
\begin{equation}
E_{e,t}
=
\begin{cases}
0,
& t=0,
\\[2mm]
\displaystyle
E_{e,t-1}
+
\frac{
P_{e,t}^{\mathrm{elec}}
}{
1000
}
\Delta t_h,
& t>0.
\end{cases}
\label{eq:cumulative-energy}
\end{equation}
The factor $1/1000$ converts watts to kilowatts, while
$\Delta t_h=1/60$ converts the one-minute interval to hours.

\subsection{Internal Humidity}
\label{app:internal-humidity}

Internal relative humidity describes how much moisture is present in
the air inside the refrigerated compartment, relative to the maximum
amount that the air can contain at the same temperature. It is
expressed as a percentage. Door openings can allow moist ambient air
to enter the compartment \cite{HeHumidityReview2025,IpekDincer2023},
and air leakage through a degraded door seal can also increase this
exchange \cite{LiuGasket2024}. The simulator uses these relationships
to generate an optional internal-humidity sensor output. This
subsection first introduces the three quantities used in the
calculation, then shows how the target humidity is set, how the
internal humidity moves toward it minute by minute, and finally how
the released sensor output is obtained.

\paragraph{Humidity quantities:}
The humidity calculation uses three related quantities, which it is
useful to distinguish before looking at the equations. The target
humidity $H_{e,t}^{\mathrm{target}}$ is the value that the current
door and seal conditions point toward at each minute, but it is not
itself the humidity inside the compartment. The internal humidity
$\bar H_{e,t}^{\mathrm{int}}$ is the simulator's own estimate of the
actual humidity inside the compartment, and it moves gradually toward
the target. Finally, $H_{e,t}^{\mathrm{int}}$
is the sensor output, obtained
by adding a small random variation to $\bar H_{e,t}^{\mathrm{int}}$.

\paragraph{Target humidity:}
At each operating minute, the simulator first calculates the target
humidity from the door-seal efficiency et $\eta_{e,t}^{\mathrm{seal}}$ $\in [0,1]$ and the door state $s_{e,t} \in \{0,1\}$, both
defined in Section~\ref{app:door-use} of the Appendix,
\begin{equation}
H_{e,t}^{\mathrm{target}}
=
50
+
20
\left(
1-\eta_{e,t}^{\mathrm{seal}}
\right)
+
5s_{e,t}^{\mathrm{door}}.
\label{eq:internal-humidity-target}
\end{equation}
The reference value of $50\%$ corresponds to a closed door with a
fully efficient seal. The seal-degradation term increases the target
humidity as $\eta_{e,t}^{\mathrm{seal}}$ decreases and can add up to
6$\%$. The door-opening term adds 5$\%$ 
when the door is open. The target humidity therefore ranges from
$50\%$ to $61\%$. The coefficients 20 and 5 were chosen to give seal
degradation and door opening moderate effects on internal humidity.

\paragraph{Internal humidity update:}
The internal humidity does not immediately take the target value.
Instead, it moves gradually toward it, one small step per minute. At
each minute, the simulator adds $1\%$ of the difference between the
current target and the previous internal value,
\begin{equation}  
\bar H_{e,t}^{\mathrm{int}} = 
\begin{cases}
   \bar H_{e,t-1}^{\mathrm{int}} + 0.01 \left( H_{e,t}^{\mathrm{target}} - \bar H_{e,t-1}^{\mathrm{int}} \right) & \text{if } t > 0 \\
   45 & \text{if } t = 0
\end{cases}
\label{eq:internal-humidity-state}
\end{equation}

The intial internal humidity is $45\%$ and the
target is $50\%$, the difference is 5$\%$, and the
internal value increases by $0.01\times5=0.05$ $\%$, to
$45.05\%$. 

Once the internal value is updated for the current minute, the
released sensor output is obtained by adding a small random
variation to it, so that the generated data does not look perfectly
smooth,
\begin{equation}
\varepsilon_{e,t}^{H}
\sim
\mathcal N
\left(
0,
(\sigma^{H})^2
\right),
\qquad
\sigma^{H}
=
1\text{ $\%$}.
\label{eq:internal-humidity-noise}
\end{equation}
The resulting value is then limited to stay within the interval
$[20\%,95\%]$, so that random variation alone cannot produce an
unrealistic humidity value,
\begin{equation}
H_{e,t}^{\mathrm{int}}
=
\min\!\left(
95,\,
\max\!\left(
20,\,
\bar H_{e,t}^{\mathrm{int}}
+
\varepsilon_{e,t}^{H}
\right)
\right).
\label{eq:internal-humidity-output}
\end{equation}

\subsection{Frost Thickness}
\label{app:frost-thickness}

Frost forms on the evaporator when moisture freezes on its cold
surface. Door openings increase moisture entry and can therefore
increase frost formation. During defrost, the evaporator is heated to
remove accumulated frost \cite{IpekDincer2024}. The simulator uses
these relationships to generate an optional frost-thickness sensor
output, enabled through \(\theta_e\).

The internal frost thickness
\(\bar B_{e,t}^{\mathrm{frost}}\), in millimetres, is a component of
\(Z_{e,t}\). Its update uses the compressor state
\(s_{e,t}^{\mathrm{comp}}\) (see \ref{app:thermostat-compressor} in the Appendix), the door state
\(s_{e,t}^{\mathrm{door}}\) (see \ref{app:door-use} in the Appendix), and the
defrost state \(s_{e,t}^{\mathrm{def}}\) and heater efficiency
\(\eta_{e,t}^{\mathrm{def}}\) (see \ref{app:defrost-state} in the Appendix).

Frost grows when the compressor is active and defrost is inactive.
Door opening increases this growth, while an active defrost cycle
removes accumulated frost. The internal frost thickness is
\begin{equation}
\bar B_{e,t}^{\mathrm{frost}}
=
\begin{cases}
0,
& t=0,
\\[2mm]
\max\!\left[
0,\,
\bar B_{e,t-1}^{\mathrm{frost}}
+
0.001\left(2-\eta_{e,t}^{\mathrm{def}}\right)
+
0.005
\right],
&
s_{e,t}^{\mathrm{comp}}=1,\;
s_{e,t}^{\mathrm{def}}=0,\;
s_{e,t}^{\mathrm{door}}=1,
\\[2mm]
\max\!\left[
0,\,
\bar B_{e,t-1}^{\mathrm{frost}}
+
0.001\left(2-\eta_{e,t}^{\mathrm{def}}\right)
\right],
&
s_{e,t}^{\mathrm{comp}}=1,\;
s_{e,t}^{\mathrm{def}}=0,\;
s_{e,t}^{\mathrm{door}}=0,
\\[2mm]
\max\!\left[
0,\,
\bar B_{e,t-1}^{\mathrm{frost}}
-
0.15\eta_{e,t}^{\mathrm{def}}
\right],
&
s_{e,t}^{\mathrm{def}}=1,
\\[2mm]
\bar B_{e,t-1}^{\mathrm{frost}},
&
\text{otherwise}.
\end{cases}
\label{eq:frost-thickness-update}
\end{equation}

The coefficient \(0.001\,\mathrm{mm}\) gives gradual frost accumulation
during normal cooling. An open door adds \(0.005\,\mathrm{mm}\) per
minute, giving moisture entry a stronger effect on frost growth.
During defrost, the larger coefficient \(0.15\,\mathrm{mm}\) allows
frost to be removed much faster than it accumulates. A lower
\(\eta_{e,t}^{\mathrm{def}}\) increases frost accumulation and reduces
frost removal, making defrost-heater degradation visible in this
output. The maximum with zero prevents negative frost thickness.

The frost-thickness sensor output
\(B_{e,t}^{\mathrm{frost}}\) is obtained by rounding the internal value
to two decimal places,
\begin{equation}
B_{e,t}^{\mathrm{frost}}
=
\operatorname{round}
\left(
\bar B_{e,t}^{\mathrm{frost}},
2
\right).
\label{eq:frost-thickness-output}
\end{equation}

\subsection{Vibration and Acoustic Noise}
\label{app:vibration-noise}

Vibration describes the mechanical motion produced by refrigerator
components, while acoustic noise describes the sound produced during
operation. The simulator generates the vibration output
\(V_{e,t}^{\mathrm{vib}}\) and the acoustic-noise output
\(N_{e,t}^{\mathrm{ac}}\) only when the corresponding sensor outputs
are enabled in \(\theta_e\).

\paragraph{Vibration:}
The compressor and fan are the main sources of vibration in
refrigerators \cite{Nunes2026,Zarate2021}. Compressor condition can
also affect measured vibration \cite{Solmaz2023}. The simulator
therefore changes the vibration level according to compressor
operation, compressor degradation, and fan degradation.

Vibration is expressed in units of standard gravitational acceleration,
denoted by \(\mathrm{g}\), where
\(1\,\mathrm{g}\approx9.81\,\mathrm{m\,s^{-2}}\). Using the fan
efficiency \(\eta_{e,t}^{\mathrm{fan}}\) and fan speed \(RPM_{e,t}\)
defined in Sections~\ref{app:thermal-behaviour} and
\ref{app:electrical-quantities} of the Appendix, respectively, the fan contribution is
\begin{equation}
V_{e,t}^{\mathrm{fan}}
=
\begin{cases}
0.10
\left(
1-\eta_{e,t}^{\mathrm{fan}}
\right),
&
\eta_{e,t}^{\mathrm{fan}}<1,\;
RPM_{e,t}>0,
\\[2mm]
0,
& \text{otherwise}.
\end{cases}
\label{eq:fan-vibration-contribution}
\end{equation}

The coefficient 0.10 gives a gradual increase in vibration as fan
efficiency decreases while keeping the fan contribution smaller than
the main compressor contribution.

A small random variation is added to avoid perfectly regular vibration
values,
\begin{equation}
\varepsilon_{e,t}^{V}
\sim
\mathcal N\!\left(
0,
(\sigma^{V})^2
\right).
\label{eq:vibration-perturbation}
\end{equation}
where
\[
\sigma^{V}=0.02\,\mathrm{g}.
\]
This value keeps the random variation small relative to the vibration
produced when the compressor is active.

Using the compressor state \(s_{e,t}^{\mathrm{comp}}\), mechanical-wear
factor \(w_{e,t}^{\mathrm{mec}}\), compressor efficiency
\(\eta_{e,t}^{\mathrm{comp}}\), and fan contribution
\(V_{e,t}^{\mathrm{fan}}\), the vibration output is
\begin{equation}
V_{e,t}^{\mathrm{vib}}
=
\max\!\left[
0,\,
0.02
+
s_{e,t}^{\mathrm{comp}}
\left\{
0.18
+
0.25w_{e,t}^{\mathrm{mec}}
+
0.15
\left(
1-\eta_{e,t}^{\mathrm{comp}}
\right)
\right\}
+
V_{e,t}^{\mathrm{fan}}
+
\varepsilon_{e,t}^{V}
\right].
\label{eq:vibration-output}
\end{equation}

The value \(0.02\,\mathrm{g}\) provides a small vibration level when
the compressor is inactive, while \(0.18\,\mathrm{g}\) makes
compressor operation the main contribution. The coefficients 0.25
and 0.15 allow mechanical wear and efficiency loss to progressively
increase vibration before failure. These coefficients are fixed for
data generation.

\paragraph{Acoustic noise:}
The compressor and fan are important sources of refrigerator noise
\cite{Zarate2021,LiNoise2025}. Compressor sound also changes with
operating conditions and rotational speed \cite{Catak2024}. The
simulator therefore calculates separate sound levels according to the
components that are operating.

For refrigerators using the modulated configuration
\(b_e^{\mathrm{mod}}=1\), the simulator applies the acoustic scaling
factor
\begin{equation}
\gamma_e^{\mathrm{ac}}
=
\begin{cases}
0.7,
& b_e^{\mathrm{mod}}=1,\\
1,
& b_e^{\mathrm{mod}}=0.
\end{cases}
\label{eq:acoustic-scaling-factor}
\end{equation}

The value 0.7 gives the modulated configuration a lower acoustic level
while preserving changes caused by operating conditions. It is fixed
for data generation.

The base and door-open sound levels are
\begin{equation}
N^{\mathrm{base}}
=
35,
\qquad
N^{\mathrm{door}}
=
45.
\label{eq:background-door-acoustic-levels}
\end{equation}
Both values are expressed in decibels. The higher door-open value
makes this operating condition distinguishable from the base acoustic
level.

The compressor is one of the main noise sources in household
refrigerators \cite{Zarate2021,Catak2024}. Using the mechanical-wear
factor \(w_{e,t}^{\mathrm{mec}}\), compressor efficiency
\(\eta_{e,t}^{\mathrm{comp}}\), and condenser thermal resistance
\(R_{e,t}^{\mathrm{th}}\), its sound level is
\begin{equation}
N_{e,t}^{\mathrm{comp}}
=
\gamma_e^{\mathrm{ac}}
\left[
42
+
8w_{e,t}^{\mathrm{mec}}
+
5
\left(
1-\eta_{e,t}^{\mathrm{comp}}
\right)
+
3
\left(
R_{e,t}^{\mathrm{th}}-1
\right)
\right].
\label{eq:compressor-acoustic-candidate}
\end{equation}

The reference value of \(42\,\mathrm{dB}\) places an operating
compressor above the base sound level. The coefficients 8, 5, and 3
give mechanical wear the strongest degradation effect, followed by
compressor-efficiency loss and increased condenser thermal resistance.
The exact values are fixed for data generation.

Fan operation also contributes to refrigerator noise
\cite{Zarate2021,LiNoise2025}. The fan-speed scale
\(\Delta RPM^{\mathrm{ac}}\), expressed in rpm, is
\begin{equation}
\Delta RPM^{\mathrm{ac}}
=
800.
\label{eq:acoustic-rpm-scale}
\end{equation}
This scale keeps small fan-speed changes from producing large acoustic
changes.

Using the fan speed \(RPM_{e,t}\) and fan efficiency
\(\eta_{e,t}^{\mathrm{fan}}\), the fan sound level is
\begin{equation}
N_{e,t}^{\mathrm{fan}}
=
\gamma_e^{\mathrm{ac}}
\left[
38
+
4
\frac{
RPM_{e,t}
-
RPM_e^{\mathrm{base}}
}{
\Delta RPM^{\mathrm{ac}}
}
+
6
\left(
1-\eta_{e,t}^{\mathrm{fan}}
\right)
\right].
\label{eq:fan-acoustic-candidate}
\end{equation}

The reference value \(38\,\mathrm{dB}\) keeps the fan above the base
sound level but below the reference compressor level. Fan-speed
changes modify this level gradually, while the last term increases
noise as fan efficiency decreases.

During defrost, the sound level is
\begin{equation}
N_e^{\mathrm{def}}
=
40\gamma_e^{\mathrm{ac}}.
\label{eq:defrost-acoustic-candidate}
\end{equation}
The value \(40\,\mathrm{dB}\) places the defrost condition above the
base level and within the range of the other simulated operating
sources. This value is fixed for data generation.

\paragraph{Sound level:}
Different refrigerator components can contribute to noise depending
on the current operating condition
\cite{Zarate2021,LiNoise2025}. According to the compressor state
\(s_{e,t}^{\mathrm{comp}}\), fan speed \(RPM_{e,t}\), defrost state
\(s_{e,t}^{\mathrm{def}}\), and door state
\(s_{e,t}^{\mathrm{door}}\), the simulator forms the set
\begin{equation}
\begin{aligned}
\mathcal S_{e,t}^{\mathrm{ac}}
={}&
\{N^{\mathrm{base}}\}
\\
&{}\cup
\begin{cases}
\{N_{e,t}^{\mathrm{comp}}\},
& s_{e,t}^{\mathrm{comp}}=1,\\
\varnothing,
& \text{otherwise},
\end{cases}
\\
&{}\cup
\begin{cases}
\{N_{e,t}^{\mathrm{fan}}\},
& RPM_{e,t}>0,\\
\varnothing,
& \text{otherwise},
\end{cases}
\\
&{}\cup
\begin{cases}
\{N_e^{\mathrm{def}}\},
& s_{e,t}^{\mathrm{def}}=1,\\
\varnothing,
& \text{otherwise},
\end{cases}
\\
&{}\cup
\begin{cases}
\{N^{\mathrm{door}}\},
& s_{e,t}^{\mathrm{door}}=1,\\
\varnothing,
& \text{otherwise}.
\end{cases}
\end{aligned}
\label{eq:active-acoustic-candidates}
\end{equation}

Because the simulator is intended to identify the dominant operating
source rather than reproduce the physical combination of multiple
sound sources, it keeps the largest level,
\begin{equation}
N_{e,t}^{\mathrm{dom}}
=
\max
\left\{
\mathcal S_{e,t}^{\mathrm{ac}}
\right\}.
\label{eq:dominant-acoustic-candidate}
\end{equation}

Finally, a small random variation is generated,
\begin{equation}
\varepsilon_{e,t}^{N}
\sim
\mathcal N\!\left(
0,
(\sigma^{N})^2
\right).
\label{eq:acoustic-noise-perturbation}
\end{equation}
where
\[
\sigma^{N}=1.5\,\mathrm{dB}.
\]
This value introduces small minute-to-minute changes without
dominating the differences between operating conditions.

The acoustic-noise output is then
\begin{equation}
N_{e,t}^{\mathrm{ac}}
=
N_{e,t}^{\mathrm{dom}}
+
\varepsilon_{e,t}^{N}.
\label{eq:acoustic-noise-output}
\end{equation}

\subsection{Pressures and Estimated COP}
\label{app:pressures-cop}

Suction pressure is the refrigerant pressure where it enters the
compressor, while discharge pressure is the pressure where it leaves
the compressor.
Both pressures depend on compressor operation and refrigeration
conditions \cite{PardoCely2023,IpekDincer2024}. The simulator generates
the optional outputs \(P_{e,t}^{\mathrm{suc}}\) and
\(P_{e,t}^{\mathrm{dis}}\), in bar. It also generates the optional
estimated coefficient of performance \(COP_{e,t}\), which describes
refrigerator cooling efficiency.

\paragraph{Suction pressure:}
The suction pressure $P_{e,t}^{\mathrm{suc}}$ changes according to compressor operation and
compressor condition. A small random variation is included, with a
standard deviation of \(0.05\,\mathrm{bar}\) when the compressor is
inactive and \(0.10\,\mathrm{bar}\) when it is active,
\begin{align}
\varepsilon_{e,t}^{\mathrm{suc,off}}
&\sim
\mathcal N\!\left(
0,
(\sigma^{\mathrm{suc,off}})^2
\right),
&
\varepsilon_{e,t}^{\mathrm{suc,on}}
&\sim
\mathcal N\!\left(
0,
(\sigma^{\mathrm{suc,on}})^2
\right),
\end{align}
where
\[
\sigma^{\mathrm{suc,off}}=0.05,
\qquad
\sigma^{\mathrm{suc,on}}=0.10.
\]
The larger variation during compressor operation allows slightly
greater short-term pressure changes while the refrigeration cycle is
active.

Using the compressor state \(s_{e,t}^{\mathrm{comp}}\) and the
compressor-efficiency factor \(\eta_{e,t}^{\mathrm{comp}}\), the
suction pressure is
\begin{equation}
P_{e,t}^{\mathrm{suc}}
=
\begin{cases}
1
+
\varepsilon_{e,t}^{\mathrm{suc,off}},
&
s_{e,t}^{\mathrm{comp}}=0,
\\[2mm]
\max\!\left[
0.5,\,
2.5
+
2
\left(
1-\eta_{e,t}^{\mathrm{comp}}
\right)
+
\varepsilon_{e,t}^{\mathrm{suc,on}}
\right],
&
s_{e,t}^{\mathrm{comp}}=1.
\end{cases}
\label{eq:suction-pressure}
\end{equation}

The reference levels of \(1\,\mathrm{bar}\) and
\(2.5\,\mathrm{bar}\) distinguish inactive and active compressor
conditions. The coefficient 2 makes the pressure increase
progressively as compressor efficiency decreases, while the lower
bound of \(0.5\,\mathrm{bar}\) prevents excessively low values. These
values are fixed for data generation.

\paragraph{Discharge pressure:}
Discharge pressure $P_{e,t}^{\mathrm{dis}}$ changes with compressor operation and condenser
condition. Condenser degradation can increase discharge pressure and
reduce refrigerator performance \cite{PardoCely2023}. The calculation
therefore uses the condenser thermal-resistance factor
\(R_{e,t}^{\mathrm{th}}\) and compressor efficiency
\(\eta_{e,t}^{\mathrm{comp}}\).

The discharge-pressure calculation uses a standard deviation of
\(0.10\,\mathrm{bar}\) when the compressor is inactive and
\(0.15\,\mathrm{bar}\) when it is active,
\begin{align}
\varepsilon_{e,t}^{\mathrm{dis,off}}
&\sim
\mathcal N\!\left(
0,
(\sigma^{\mathrm{dis,off}})^2
\right),
&
\varepsilon_{e,t}^{\mathrm{dis,on}}
&\sim
\mathcal N\!\left(
0,
(\sigma^{\mathrm{dis,on}})^2
\right),
\end{align}
where
\[
\sigma^{\mathrm{dis,off}}=0.10,
\qquad
\sigma^{\mathrm{dis,on}}=0.15.
\]
The larger standard deviation during compressor operation allows
greater short-term variation while the refrigeration cycle is active.

Using the compressor state \(s_{e,t}^{\mathrm{comp}}\), condenser
thermal resistance \(R_{e,t}^{\mathrm{th}}\), and compressor efficiency
\(\eta_{e,t}^{\mathrm{comp}}\), the discharge pressure is
\begin{equation}
P_{e,t}^{\mathrm{dis}}
=
\begin{cases}
6
+
\varepsilon_{e,t}^{\mathrm{dis,off}},
&
s_{e,t}^{\mathrm{comp}}=0,
\\[2mm]
\max\!\left[
5,\,
8R_{e,t}^{\mathrm{th}}
+
2
\left(
1-\eta_{e,t}^{\mathrm{comp}}
\right)
+
\varepsilon_{e,t}^{\mathrm{dis,on}}
\right],
&
s_{e,t}^{\mathrm{comp}}=1.
\end{cases}
\label{eq:discharge-pressure}
\end{equation}

The reference values of \(6\,\mathrm{bar}\) and
\(8\,\mathrm{bar}\) keep the discharge pressure above the suction
pressure and distinguish inactive and active compressor conditions.
The factor 8 makes condenser thermal resistance the main degradation
contribution to discharge pressure, consistent with the increase
reported under condenser degradation \cite{PardoCely2023}. The lower
bound of \(5\,\mathrm{bar}\) prevents excessively low discharge
pressure values.

\paragraph{Estimated COP:}
The coefficient of performance relates the cooling produced by a
refrigeration system to the energy required to produce it. Higher
values indicate better refrigeration efficiency. Fan and condenser
conditions can affect refrigerator energy consumption and COP
\cite{PardoCely2023,SunCondenser2022}.

Here, \(COP_{e,t}\) is approximated from compressor efficiency, fan
efficiency, and condenser thermal resistance rather than calculated
from refrigerant enthalpies, cooling capacity, and compressor work.

The COP estimate also includes a small random variation,
\begin{equation}
\varepsilon_{e,t}^{\mathrm{cop}}
\sim
\mathcal N\!\left(
0,
(\sigma^{\mathrm{cop}})^2
\right),
\label{eq:cop-perturbation}
\end{equation}
where
\[
\sigma^{\mathrm{cop}}=0.05.
\]
This value keeps the random variation small relative to normal changes
in COP.

Using the compressor efficiency \(\eta_{e,t}^{\mathrm{comp}}\), fan
efficiency \(\eta_{e,t}^{\mathrm{fan}}\), condenser thermal resistance
\(R_{e,t}^{\mathrm{th}}\), and compressor state
\(s_{e,t}^{\mathrm{comp}}\), the estimated COP is
\begin{equation}
COP_{e,t}
=
\begin{cases}
0,
&
s_{e,t}^{\mathrm{comp}}=0,
\\[2mm]
\max\!\left[
0.5,\,
2.5
\eta_{e,t}^{\mathrm{comp}}
\eta_{e,t}^{\mathrm{fan}}
\left[
1
-
0.5
\left(
R_{e,t}^{\mathrm{th}}-1
\right)
\right]
+
\varepsilon_{e,t}^{\mathrm{cop}}
\right],
&
s_{e,t}^{\mathrm{comp}}=1.
\end{cases}
\label{eq:estimated-cop}
\end{equation}

The reference COP of 2.5 is consistent with an experimental value
reported for an R600a domestic refrigerator
\cite{Senthilkumar2026COP}. Compressor- and fan-efficiency losses
reduce this value directly, while the coefficient 0.5 gives increasing
condenser thermal resistance a gradual negative effect. The minimum
value of 0.5 prevents the estimated COP from reaching unrealistic
negative or near-zero values. When the compressor is inactive,
\(COP_{e,t}=0\) indicates that COP is not estimated during that minute.

\section{Output Assembly}
\label{app:output-assembly}

The simulator assembles each released time-series observation from the
quantities defined above. Table~\ref{tab:sensor-output-assembly} gives
the correspondence between these quantities and the sensor-output
fields stored in the released time-series files, together with their
units and availability. In the availability column, \emph{Common}
indicates a field present in every released time-series file. For
optional sensor outputs, the entry gives the configuration flag that
controls whether the field is included.

\begingroup
\scriptsize
\setlength{\tabcolsep}{3pt}

\begin{longtable}{
>{\raggedright\arraybackslash}p{0.19\linewidth}
>{\centering\arraybackslash}p{0.15\linewidth}
>{\centering\arraybackslash}p{0.10\linewidth}
>{\raggedright\arraybackslash}p{0.31\linewidth}
>{\raggedright\arraybackslash}p{0.17\linewidth}}

\caption{Correspondence between simulator quantities and released
sensor-output fields.}
\label{tab:sensor-output-assembly}\\
\toprule
Sensor output &
Simulator quantity &
Unit/type &
Released field &
Availability \\
\midrule
\endfirsthead

\toprule
Sensor output &
Simulator quantity &
Unit/type &
Released field &
Availability \\
\midrule
\endhead

\bottomrule
\endfoot

Fridge temperature &
\(T_{e,t}^{\mathrm{fr}}\) &
\({}^{\circ}\mathrm{C}\) &
\cfgfield{temp_fridge_c} &
Common \\

Freezer temperature &
\(T_{e,t}^{\mathrm{fz}}\) &
\({}^{\circ}\mathrm{C}\) &
\cfgfield{temp_freezer_c} &
Common \\

Ambient temperature &
\(T_{e,t}^{\mathrm{amb}}\) &
\({}^{\circ}\mathrm{C}\) &
\cfgfield{ambient_temp_c} &
Common \\

Evaporator temperature &
\(T_{e,t}^{\mathrm{ev}}\) &
\({}^{\circ}\mathrm{C}\) &
\cfgfield{evaporator_temp_c} &
Common \\

Condenser temperature &
\(T_{e,t}^{\mathrm{cd}}\) &
\({}^{\circ}\mathrm{C}\) &
\cfgfield{condenser_temp_c} &
Common \\

Compressor state &
\(s_{e,t}^{\mathrm{comp}}\) &
binary &
\cfgfield{compressor_state} &
Common \\

Compressor current &
\(I_{e,t}^{\mathrm{comp}}\) &
A &
\cfgfield{compressor_current_a} &
Common \\

Duty cycle &
\(DC_{e,t}\) &
\% &
\cfgfield{duty_cycle} &
Common \\

Defrost state &
\(s_{e,t}^{\mathrm{def}}\) &
binary &
\cfgfield{defrost_state} &
Common \\

Defrost-heater current &
\(I_{e,t}^{\mathrm{def}}\) &
A &
\cfgfield{defrost_heater_current_a} &
Common \\

Fan speed &
\(RPM_{e,t}\) &
rpm &
\cfgfield{fan_rpm} &
Common \\

Door state &
\(s_{e,t}^{\mathrm{door}}\) &
binary &
\cfgfield{door_open} &
Common \\

Door-opening duration &
\(L_{e,t}^{\mathrm{door}}\) &
min &
\cfgfield{door_open_duration} &
Common \\

Electrical power &
\(P_{e,t}^{\mathrm{elec}}\) &
W &
\cfgfield{power_w} &
Common \\

Cumulative energy &
\(E_{e,t}\) &
kWh &
\cfgfield{energy_kwh} &
Common \\

Acoustic noise &
\(N_{e,t}^{\mathrm{ac}}\) &
dB &
\cfgfield{noise_db} &
\cfgfield{has_noise_db} \\

Vibration &
\(V_{e,t}^{\mathrm{vib}}\) &
\(\mathrm{g}\) &
\cfgfield{vibration_g} &
\cfgfield{has_vibration} \\

Internal humidity &
\(H_{e,t}^{\mathrm{int}}\) &
\% &
\cfgfield{humidity_percent} &
\cfgfield{has_humidity} \\

Frost thickness &
\(B_{e,t}^{\mathrm{frost}}\) &
mm &
\cfgfield{frost_thickness_mm} &
\cfgfield{has_frost_thickness} \\

Suction pressure &
\(P_{e,t}^{\mathrm{suc}}\) &
bar &
\cfgfield{suction_pressure_bar} &
\cfgfield{has_suction_pressure} \\

Discharge pressure &
\(P_{e,t}^{\mathrm{dis}}\) &
bar &
\cfgfield{discharge_pressure_bar} &
\cfgfield{has_discharge_pressure} \\

Estimated COP &
\(COP_{e,t}\) &
dimensionless &
\cfgfield{cop_estimate} &
\cfgfield{has_cop} \\

Door-opening count &
\(C_{e,t}^{\mathrm{door}}\) &
count &
\cfgfield{door_open_count} &
\cfgfield{has_door_count} \\

Ambient humidity &
\(H_{e,t}^{\mathrm{amb}}\) &
\% &
\cfgfield{ambient_humidity_percent} &
\cfgfield{has_ambient_humidity} \\

\end{longtable}
\endgroup

Before each time-series observation is generated, fridge, freezer,
ambient, evaporator, and condenser temperatures, compressor and
defrost-heater currents, and frost thickness are rounded to two decimal
places. Duty cycle, electrical power, acoustic noise, internal humidity,
and ambient humidity are rounded to one decimal place. Cumulative
energy, vibration, suction pressure, discharge pressure, and estimated
COP are rounded to three decimal places. Fan speed is rounded to the
nearest whole rpm. Binary states, door-opening duration, and count
variables are stored as integers. These rounding operations are applied
only to the released values and do not modify \(Z_{e,t}\).

The fridge- and freezer-temperature fields remain in the common output
structure when the corresponding compartment is absent. In that case,
the released value is zero and serves only as a structural placeholder,
not as a temperature measurement. Optional sensor-output fields are
omitted when their corresponding availability flag is disabled.

In addition to the sensor-output vector \(X_t^{(e)}\), each observation
at instant \(t\) contains the fields listed in
Table~\ref{tab:row-annotation-fields}.

\begin{table}[H]
\centering
\small
\caption{Additional fields stored with each time-series observation.}
\label{tab:row-annotation-fields}

\begin{tabular}{
>{\raggedright\arraybackslash}p{0.31\linewidth}
>{\centering\arraybackslash}p{0.13\linewidth}
>{\raggedright\arraybackslash}p{0.48\linewidth}}

\toprule
Released field &
Unit/type &
Meaning \\
\midrule

\cfgfield{timestamp} &
datetime &
Timestamp corresponding to instant \(t\). \\

\cfgfield{fault_indicator} &
binary &
Equipment-state label \(y_{e,t}\) defined in
Eq.~\eqref{eq:equipment-state-label}. \\

\cfgfield{degradation_indicator} &
binary &
Equal to 1 when
\(t_{e,j}^{\mathrm{deg}}\leq t<t_{e,j}^{\mathrm{fail}}\)
for a failure \(j\), and 0 otherwise. \\

\cfgfield{fault_count} &
count &
Cumulative number of failure intervals that have started by instant
\(t\). \\

\cfgfield{time_since_last_fault_min} &
min &
Number of minutes since the most recently completed failure interval,
with value \(-1\) before the first completed failure. \\

\bottomrule
\end{tabular}
\end{table}

\end{document}